\documentclass[lettersize,journal]{IEEEtran}
\usepackage{amsmath,amsfonts}
\usepackage{algorithmic}
\usepackage{algorithm}
\usepackage{array}
\usepackage[caption=false,font=normalsize,labelfont=sf,textfont=sf]{subfig}
\usepackage{textcomp}
\usepackage{stfloats}
\usepackage{url}
\usepackage{verbatim}
\usepackage{graphicx}
\usepackage{cite}
\usepackage[hidelinks,bookmarks=false]{hyperref}
\usepackage{multirow}

\hypersetup{hypertex=true,
	colorlinks=true,
	linkcolor=blue,
	anchorcolor=blue,
	citecolor=blue}
\newtheorem{theorem}{\textbf{Theorem}}
\newtheorem{lemma}{\textbf{Lemma}}
\newtheorem{definition}{\textbf{Definition}}
\newtheorem{remark}{\textbf{Remark}}

\def\BibTeX{{\rm B\kern-.05em{\sc i\kern-.025em b}\kern-.08em
		T\kern-.1667em\lower.7ex\hbox{E}\kern-.125emX}}

\newcommand{\URE}{\textsc{URE}}

\begin{document}
	\title{Cost-Sensitive Unbiased Risk Estimation for Multi-Class Positive-Unlabeled Learning}
	\author{Miao Zhang, Junpeng Li, \emph{Member}, \emph{IEEE}, Changchun Hua, \emph{Fellow}, \emph{IEEE} and Yana Yang, \emph{Member}, \emph{IEEE} \thanks{M. Zhang, J. Li, C. Hua, and Y. Yang are with the Engineering Research Center of the Ministry of Education for Intelligent Control System and Intelligent Equipment, Yanshan University, Qinhuangdao, China (zhangmiao@stumail.ysu.edu.cn; jpl@ysu.edu.cn; cch@ysu.edu.cn; yyn@ysu.edu.cn).}}
	\date{}
	\maketitle
	
	\begin{abstract}
		Positive--Unlabeled (PU) learning considers settings in which only positive and unlabeled data are available, while negatives are missing or left unlabeled. This situation is common in real applications where annotating reliable negatives is difficult or costly. Despite substantial progress in PU learning, the multi-class case (MPU) remains challenging: many existing approaches do not ensure \emph{unbiased risk estimation}, which limits performance and stability. We propose a cost-sensitive multi-class PU method based on \emph{adaptive loss weighting}. Within the empirical risk minimization framework, we assign distinct, data-dependent weights to the positive and \emph{inferred-negative} (from the unlabeled mixture) loss components so that the resulting empirical objective is an unbiased estimator of the target risk. We formalize the MPU data-generating process and establish a generalization error bound for the proposed estimator. Extensive experiments on \textbf{eight} public datasets, spanning varying class priors and numbers of classes, show consistent gains over strong baselines in both accuracy and stability.
	\end{abstract}

	\section{Introduction}
	Supervised learning typically assumes access to fully labeled datasets—a requirement that can be prohibitively expensive in domains such as medical imaging, fault diagnosis, or content moderation, where negative or rare-class samples are costly to curate and verify~\cite{zhou2018brief, 10847040, 10039501}. To mitigate annotation burden, weakly supervised learning leverages incompletely or imprecisely labeled data and has grown into a broad family of methods~\cite{9444588, 10172241, 10472151}. Among them, PU learning considers scenarios where only a subset of positive examples are labeled, while the remaining data are kept unlabeled; crucially, the unlabeled pool is a mixture of positives and negatives~\cite{bekker2020learning, jiang2023positive, 10870373, chang2021positive}. PU learning has proven effective in binary settings such as fault detection and medical diagnosis, where obtaining verified negatives is impractical~\cite{9583860, 10792933}.
	
	In many real-world multi-class problems, it is unrealistic to obtain labeled examples for all classes during training. This motivates the MPU setting~\cite{ijcai2017p444, shu2020learning}, where training data comprise a labeled set and an unlabeled set. The labeled set contains true labels only for a subset of classes—referred to as the \emph{observed classes}—whereas the unlabeled set is drawn from the mixture over \emph{all} classes, including both observed and \emph{unobserved} classes, but without labels. In MPU, the notion of a “positive” example is class-relative: under a one-vs-rest view, instances belonging to a target class are positives for that class, and instances from any other class act as negatives for that class.
	
	In the absence of additional structure or side information, individual unobserved classes cannot be uniquely identified from the unlabeled mixture over all classes. Reflecting this practical limitation, we target the immediately useful goal of \emph{reliably separating the observed classes from the unlabeled pool} under scarce labels. Concretely, our objective is robust one-vs-rest (OVR) detection for each observed class; we do not require the unlabeled remainder to be partitioned into its constituent unobserved classes. This focus aligns with deployment scenarios in, for example, astronomical discovery streams~\cite{fotopoulou2024review} and e-commerce review mining~\cite{bhat2017identifying}, where practitioners need to surface known categories of interest while suppressing everything else.
	
	Despite steady progress, existing MPU methods face two persistent challenges. First, many approaches inherit bias from risk estimators or rely on specialized losses whose constant-sum mismatch introduces non-vanishing constants in generalization analyses, weakening theoretical guarantees~\cite{shu2020learning}. Second, optimization can be brittle in the presence of heavily imbalanced class priors and entirely unlabeled classes; pseudo-labeling pipelines may amplify early mistakes and get trapped in local optima~\cite{ijcai2017p444}. Techniques developed for binary PU, such as non-negative risk estimators~\cite{kiryo2017positive}, do not directly address multi-class intricacies or the need to calibrate decisions across several OVR tasks.
	
	We introduce \emph{CSMPU}, a framework tailored to observed-class detection under MPU. CSMPU formulates a per-class risk by combining a cost-sensitive OVR loss with a non-negativity correction that stabilizes training on unlabeled data. The cost-sensitive design places appropriate emphasis on minority observed classes and enables explicit control of each class's contribution, while the correction term prevents pathological solutions driven by the unlabeled mixture. The result is a simple, modular objective that integrates cleanly with modern neural encoders and requires no additional supervision beyond the existing labeled positives.
	
	\begin{itemize}
		\item \textbf{Problem focus.} We formalize MPU as an \emph{observed-class detection} task: for each observed class, separate its positives from the unlabeled mixture over all classes, without attempting to disentangle the remainder into specific unobserved classes.
		\item \textbf{Cost-sensitive OVR learning.} We design a practical objective that accounts for class imbalance and yields calibrated decisions for each observed class, while remaining stable in the presence of unlabeled data.
		\item \textbf{Theoretical and empirical support.} We provide learning guarantees based on standard capacity measures and show consistent gains over MPU baselines in both accuracy and stability across varying class priors and numbers of classes.
	\end{itemize}
	
	Section~\ref{sec:2} reviews multi-class supervised learning, PU learning, and cost-sensitive learning.
	Section~\ref{sec:3} formalizes the MPU problem and presents the CSMPU objective.
	Section~\ref{sec:4} provides generalization analyses.
	Section~\ref{sec:corrected} introduces the non-negativity correction (also referred to as the Corrected Function).
	Section~\ref{sec:5} reports experiments and ablations, and Section~\ref{sec:6} concludes.
	\section{PRELIMINARIES}\label{sec:2}
	\subsection{Ordinary Multi-class Supervised Learning}
	Let the feature space be $\mathcal{X} \subseteq \mathbb{R}^d$ and the label space be $\mathcal{Y}=\{1,\dots,k\}$, representing a $k$-class classification problem.
	Assume that sample pairs $(\boldsymbol{x},y)$ are drawn i.i.d.\ from the joint distribution $p(\boldsymbol{x},y)$, where $\boldsymbol{x}\in\mathcal{X}$ and $y\in\mathcal{Y}$.
	The goal is to learn a classifier $f:\mathcal{X}\to\mathbb{R}^k$ that minimizes the expected risk
	\begin{equation*}
		\mathcal{R}(f)\;=\;\mathbb{E}_{(\boldsymbol{x},y)\sim p(\boldsymbol{x},y)}\!\left[\mathcal{L}\big(f(\boldsymbol{x}),y\big)\right],
	\end{equation*}
	where $\mathcal{L}:\mathbb{R}^k\times\mathcal{Y}\to\mathbb{R}$ is a multi-class loss function.
	The learning objective is therefore to find the optimal classifier
	\begin{equation*}
		f^\ast \in \arg\min_{f\in\mathcal{F}} \mathcal{R}(f),
	\end{equation*}
	where $\mathcal{F}$ denotes the hypothesis class of candidate classifiers.
	Given a sample $\boldsymbol{x}$, the predicted label is
	\begin{equation*}
		\hat{y}\;=\;\arg\max_{y\in\mathcal{Y}} f_y(\boldsymbol{x}),
	\end{equation*}
	where $f_y(\boldsymbol{x})$ denotes the $y$-th component of $f(\boldsymbol{x})\in\mathbb{R}^k$.
	\subsection{Positive and Unlabeled Learning}
	In many real-world scenarios, negative class samples are either difficult to obtain or remain unlabeled, rendering conventional supervised classification methods inapplicable~\cite{bekker2020learning,elkan2008learning}. To address this challenge, PU learning provides a framework for training classifiers using only positive and unlabeled samples~\cite{bekker2020learning,jiang2023positive,chang2021positive}. The core idea of PU learning is to reconstruct the classification risk by modeling the relationship between positive and unlabeled data: specifically, the traditional joint risk over positives and negatives is rewritten into an equivalent form that depends solely on PU samples by leveraging the class prior and the unlabeled-data distribution~\cite{du2014analysis,elkan2008learning}. This reformulation underpins unbiased (or bias-corrected non-negative) risk estimators that enable effective learning without explicit negative labels~\cite{du2014analysis,kiryo2017positive}.
	\subsection{Cost-Sensitive Learning}
	In the presence of class imbalance, conventional classifiers that optimize overall accuracy tend to favor the majority class, yielding poor recognition of minority classes. To mitigate this issue, \emph{cost-sensitive learning}~\cite{elkan2001foundations, sun2007cost, zhou2010multi} provides a principled alternative in which different misclassification errors are assigned distinct weights to reflect their real-world consequences.
	
	Concretely, cost-sensitive learning specifies a $k\times k$ \emph{cost matrix} $C=[C_{ij}]$, where $C_{ij}$ denotes the penalty for predicting class $j$ when the true class is $i$ (typically $C_{ii}=0$). Training then shifts from minimizing the raw error rate to minimizing the \emph{expected misclassification cost} under the data distribution.
	
	For an input $(\boldsymbol{x},y)$ with $y\in\{1,\dots,k\}$, let $\boldsymbol{o}=(o_1,\dots,o_k)$ denote the model outputs (e.g., calibrated scores or predicted probabilities) and let $\boldsymbol{t}=(t_1,\dots,t_k)$ be the one-hot encoding of the ground-truth label. A standard formulation minimizes
	\[
	\mathbb{E}\!\left[\sum_{j=1}^k C_{y j}\, o_j\right],
	\]
	when $\boldsymbol{o}$ are class-posterior estimates, or uses a surrogate objective that upper-bounds the discrete cost $\mathbb{E}\!\big[C_{y,\hat y}\big]$ with $\hat y=\arg\max_j o_j$. This cost-sensitive design explicitly reweights errors, improving minority-class recognition without altering the label space.
	
	A cost-sensitive variant of the mean squared error (MSE) loss~\cite{khan2017cost,o2008cost} can be written as
	\begin{equation*}
		\mathcal{L}(\boldsymbol{o},\boldsymbol{t};\boldsymbol{C})
		\;=\;\sum_{j=1}^k C_{y j}\,\big(t_j - o_j\big)^2,
	\end{equation*}
	where $C_{y j}$ weights the squared error incurred when the true class is $y$ and class $j$ is predicted. This formulation places larger penalties on high-cost mistakes (e.g., predicting a majority class for a minority-class instance), steering optimization toward improved minority-class sensitivity.
	
	The corresponding expected risk over the data distribution $p(\boldsymbol{x},y)$ is
	\begin{equation*}
		\mathcal{R}_{\text{CS-MSE}}
		\;=\;
		\mathbb{E}_{(\boldsymbol{x},y)\sim p(\boldsymbol{x},y)}
		\!\left[
		\sum_{j=1}^k C_{y j}\,\big(t_j - o_j\big)^2
		\right].
	\end{equation*}
	This cost-sensitive risk encourages the model to prioritize reducing high-cost errors, and has been found effective in imbalanced classification compared with standard (cost-insensitive) objectives~\cite{sun2007cost,zhou2010multi}.
	\section{The Proposed Approach}\label{sec:3}
	In this section, we propose \emph{Cost-Sensitive Multi-class Positive--Unlabeled learning (CSMPU)}, a framework tailored to observed-class detection under MPU. CSMPU builds on a cost-sensitive OVR loss that assigns distinct penalties to different misclassification types, enabling finer control over class-wise learning dynamics.
	
	\subsection{Data Generation Process}
	We formally describe the data-generation mechanism underlying the MPU framework. Consider a $k$-class classification problem with feature space $\mathcal{X}\subseteq\mathbb{R}^d$ and label space $\mathcal{Y}=\{1,\dots,k\}$. Samples $(\boldsymbol{x},y)$ are drawn i.i.d.\ from an unknown joint distribution $p(\boldsymbol{x},y)$ over $\mathcal{X}\times\mathcal{Y}$, which factorizes as
	\begin{equation*}
		p(\boldsymbol{x},y)\;=\;p(y)\,p(\boldsymbol{x}\mid y),
	\end{equation*}
	where $p(y)$ denotes the class-prior distribution and $p(\boldsymbol{x}\mid y)$ the class-conditional distribution.
	
	\textbf{Generation of labeled datasets.}
	Under MPU, we assume access to class-conditional datasets for the \emph{observed classes}. Specifically, for each $i\in\{1,\dots,k-1\}$, the dataset $\mathcal{X}_i$ consists of samples drawn independently from
	\begin{equation*}
		p_i(\boldsymbol{x})\;:=\;p(\boldsymbol{x}\mid y=i).
	\end{equation*}
	In other words, all samples in $\mathcal{X}_i$ belong to class $i$, so we refer to $\mathcal{X}_i$ as the labeled dataset of class $i$.
	
	\textbf{Generation of the unlabeled dataset.}
	In addition to these $k-1$ pure datasets, we observe an unlabeled dataset $\mathcal{X}_k$ (the \emph{unlabeled pool}). Each sample $(\boldsymbol{x},y)$ in this set is drawn from the joint distribution $p(\boldsymbol{x},y)$, but only the feature $\boldsymbol{x}$ is observed while the true label $y$ remains hidden. Consequently, $\mathcal{X}_k$ contains a mixture of all $k$ classes, with input marginal
	\begin{equation*}
		p_u(\boldsymbol{x})\;:=\;\sum_{i=1}^k p(y=i)\,p(\boldsymbol{x}\mid y=i).
	\end{equation*}

	Under the MPU framework, the goal is to learn a multi-class classifier $f:\mathcal{X}\to\mathbb{R}^k$ from the pure class-conditional datasets $\{\mathcal{X}_i\}_{i=1}^{k-1}$ together with the unlabeled dataset $\mathcal{X}_k$, so that $f$ accurately predicts the latent class of unseen instances.
	
	\subsection{Cost-Sensitive Risk}
	To address MPU, we build on the unbiased risk estimator (\URE) and introduce a cost-sensitive variant. In the OVR view, \URE\ estimates the target risk
	\begin{equation*}
		\begin{split}
			\mathcal{R}(f)
			&= \sum_{i=1}^{k-1} \pi_i\,
			\mathbb{E}_{\boldsymbol{x}\sim p_i}\!\left[
			\tilde{\mathcal{L}}\big(f(\boldsymbol{x}), i\big)
			\right]
			\;+\;
			\mathbb{E}_{\boldsymbol{x}\sim p_u}\!\left[
			\mathcal{L}\big(f(\boldsymbol{x}), k\big)
			\right],
		\end{split}
	\end{equation*}
	where $\pi_i$ is the class prior of class $i$, $p_i(\boldsymbol{x})=p(\boldsymbol{x}\mid y=i)$ and $p_u(\boldsymbol{x})=\sum_{j=1}^k \pi_j p(\boldsymbol{x}\mid y=j)$ denote the labeled and unlabeled marginals, respectively, and $\mathcal{L}$ is a base loss. The composite loss
	\[
	\tilde{\mathcal{L}}\big(f(\boldsymbol{x}), i\big)
	:= \mathcal{L}\big(f(\boldsymbol{x}), i\big)
	- \mathcal{L}\big(f(\boldsymbol{x}), k\big)
	\]
	subtracts the unlabeled term to remove the unobserved-negative contribution, yielding an unbiased objective in expectation under MPU. The complete proof is provided in Appendix I.
	
	Although unbiased, this subtraction introduces negative components at the sample level, which can inflate variance and lead to overfitting or numerically negative empirical risks. To mitigate these issues, we adopt a \emph{cost-sensitive} design that explicitly upweights the penalty for misclassifying labeled instances as negatives (the most critical error in MPU), thereby stabilizing learning and improving per-class calibration in the presence of class imbalance.
	
	We adopt a OVR decomposition to incorporate the cost-sensitive principle into the learning framework. Injecting the cost matrix into the OVR formulation yields the per-sample loss
	\begin{equation*}
		\mathcal{L}\big(f(\boldsymbol{x}),y\big)
		\;=\;
		\ell\!\big(f_y(\boldsymbol{x}),+1\big)
		\;+\;
		\sum_{i\in \mathcal{Y}\setminus\{y\}} C_{y i}\,\ell\!\big(f_i(\boldsymbol{x}),-1\big),
	\end{equation*}
	where $C_{y i}$ encodes the penalty for predicting class $i$ when the true class is $y$, and $\ell(\cdot,\cdot)$ is a binary surrogate (e.g., logistic/sigmoid, ramp, hinge).
	
	To sharpen the separation between each labeled (observed) class and the rest, we use class-specific losses. In the MPU setup, the index $k$ denotes the unlabeled/negative meta-class. Reflecting that misclassifying a positive from class $y$ as the negative meta-class is more severe than confusing it with another observed class, we choose a degenerate cost scheme
	\[
	C_{y k}=1,\qquad C_{y j}=0\ \ \text{for all } j\in \mathcal{Y}\setminus\{y,k\}.
	\]
	Under this choice, the OVR loss for class $i$ simplifies to
	\begin{equation*}
		\mathcal{L}_i\big(f(\boldsymbol{x})\big)
		\;=\;
		\ell\!\big(f_i(\boldsymbol{x}),+1\big)
		\;+\;
		\ell\!\big(f_k(\boldsymbol{x}),-1\big),
	\end{equation*}
	where $f_j(\boldsymbol{x})$ denotes the $j$-th component of $f(\boldsymbol{x})$. This design explicitly emphasizes ``positive vs.\ negative'' discrimination for each observed class while avoiding unnecessary penalties among observed classes.
	
	For the negative meta-class $k$, we aim to prevent confusion with the most confidently predicted observed class—which is often the most ambiguous case. Accordingly, we define the negative-class loss as
	\begin{equation*}
		\begin{split}
			\mathcal{L}_k\big(f(\boldsymbol{x})\big)
			\;=\;
			\ell\!\big(f_k(\boldsymbol{x}), +1\big)
			\;+\;
			\ell\!\big(f_{i^\ast}(\boldsymbol{x}), -1\big), \\
			\qquad
			i^\ast \;=\; \arg\max_{j\in\{1,\dots,k-1\}} f_j(\boldsymbol{x}) .
		\end{split}
	\end{equation*}
	This adaptive choice imposes stronger regularization on the dominant positive classifier whenever the input is actually negative.
	
	We require the base binary surrogate to satisfy the constant--sum (symmetry) identity
	\[
	\ell(z,+1)\;+\;\ell(-z,-1)\;\equiv\; C,
	\]
	so that unobservable terms cancel in the MPU risk. In our experiments, losses that \emph{exactly} satisfy the identity include the unhinged loss
	\(
	\ell_{\mathrm{unh}}(z) = (1 - z)/2
	\)
	with $C=1$, the \emph{probability-style} sigmoid
	\(
	\ell_{\sigma}(z) = \sigma(-\gamma z) = 1/(1+e^{\gamma z})
	\)
	with $C=1$ (note this is \emph{not} the logistic log-loss), and a smooth $\tanh$-based variant with $C=1$. By contrast, hinge and ramp satisfy the identity only \emph{after} symmetrization/clipping (denoted ``sym''), which enforces $C=1$. Likewise, the logistic log-loss (softplus) violates the constant--sum property in its raw form. Table~\ref{tab:loss_variants_fmnist_k4} reports the numerical check of $\max$ and $p_{99}$ of $\lvert \ell(z)+\ell(-z)-C\rvert$ on a dense margin grid, confirming which variants satisfy the assumption and which require symmetrization before being used in our unbiased risk.
	
	Under this construction, following standard derivations~\cite{JMLR:v23:21-0946,lu2018minimal,9878485} and assuming class priors $\{\pi_i\}_{i=1}^k$ are known or estimable (e.g., via density–ratio estimation), we obtain the \textbf{cost-sensitive unbiased risk} for MPU. The class-prior–only constant arises when the negative-class expectation is rewritten using the unlabeled mixture and the constant--sum identity
	$\ell(z,+1)+\ell(-z,-1)\equiv C$
	is applied \emph{twice} (to the positive and negative parts). After algebraic cancellation of unobservable terms, this yields a model–independent shift of $-2(1-\pi_k)C$ (with $C{=}1$ in our implementation), which does not affect ERM minimizers but is necessary for exact unbiasedness and for our numerical constant--sum diagnostics.
	
	\textbf{Population risk (class $k$).}
	\begin{align}
		\mathcal{R}_k(f)
		&:= \pi_k\,\mathbb{E}_{P_k}\!\big[\ell(z_k)\big]
		\;+\; (1-\pi_k)\,\mathbb{E}_{N_k}\!\big[\ell(-z_k)\big] \nonumber\\
		&\overset{\text{PU+sym}}{=}
		2\pi_k\,\mathbb{E}_{P_k}\!\big[\ell(z_k)\big]
		\;-\; \mathbb{E}_{U}\!\big[\ell(z_k)\big]
		\;-\; 2(1-\pi_k)C, 
		\label{eq:pop-risk-sym}
	\end{align}
	where $z_k=f_k(\boldsymbol{x})$, $\mathbb{E}_{P_k}$ and $\mathbb{E}_{N_k}$ denote expectations over $p(\boldsymbol{x}\mid y=k)$ and $p(\boldsymbol{x}\mid y\neq k)$, and $\mathbb{E}_{U}$ is with respect to the unlabeled marginal $p_u$.
	
	\textbf{Empirical counterpart.}
	\begin{align}
		\hat{\mathcal{R}}_k(f)
		&:= 2\pi_k\,\hat{\mathbb{E}}_{P_k}\!\big[\ell(z_k)\big]
		\;-\; \hat{\mathbb{E}}_{U}\!\big[\ell(z_k)\big]
		\;-\; 2(1-\pi_k)C, \label{eq:emp-risk-sym}
	\end{align}
	with sample means $\hat{\mathbb{E}}_{P_k}[\cdot]$ over positives of class $k$ and $\hat{\mathbb{E}}_{U}[\cdot]$ over unlabeled data.
	
	\begin{theorem}
		\label{thm:csmpu_unbiased_risk}
		Assume the data distribution consists of $k-1$ labeled (observed) classes with priors $\pi_i=p(y=i)$ for $i=1,\dots,k-1$, and an unlabeled mixture with density $p_u(\boldsymbol{x})=\sum_{j=1}^k \pi_j p_j(\boldsymbol{x})$. Then the CSMPU population risk is
		\begin{equation*}
			\begin{split}
				\mathcal{R}(f)
				\;&=\;
				\sum_{i=1}^{k-1} 2\pi_i \, \mathbb{E}_{\boldsymbol{x}\sim p_i}\!\big[ \mathcal{L}_i(f(\boldsymbol{x})) \big] \\
				\;&+\;
				\mathbb{E}_{\boldsymbol{x}\sim p_u}\!\big[ \mathcal{L}_k(f(\boldsymbol{x})) \big]
				\;-\; 2(1-\pi_k)C,
			\end{split}
		\end{equation*}
		and under our normalization $C{=}1$ this reduces to the constant $-2(1-\pi_k)$.
	\end{theorem}
	The completely proof is provided in Appendix II.
	\noindent
	The constant offset stems solely from the mixture decomposition $p_u(\boldsymbol{x})=\pi_k p(\boldsymbol{x}\mid y=k)+(1-\pi_k)p(\boldsymbol{x}\mid y\neq k)$ and the constant--sum $\ell(z)+\ell(-z)=C$, leaving a model–independent term that does not affect the ERM minimizer yet preserves exact unbiasedness. The corresponding empirical risk is
	\begin{equation*}
		\begin{split}
			\hat{\mathcal{R}}(f)
			\;&=\;
			\sum_{i=1}^{k-1} 2\pi_i \,\frac{1}{n_i}\sum_{j=1}^{n_i} \mathcal{L}_i\!\big(f(\boldsymbol{x}_j^{(i)})\big) \\
			\;&+\;
			\frac{1}{n_k}\sum_{j=1}^{n_k} \mathcal{L}_k\!\big(f(\boldsymbol{x}_j^{(k)})\big)
			\;-\; 2(1-\pi_k)C,
		\end{split}
	\end{equation*}
	where $\{\boldsymbol{x}_j^{(i)}\}_{j=1}^{n_i}$ are samples from class $i$.
	A complete derivation is provided in the Appendix.
	
	\medskip
	\noindent\textit{Remarks.}
	(i) With $C{=}1$ (our default for symmetric or symmetrized losses), the constant term reduces to $-2(1-\pi_k)$, consistent with the expression used throughout the paper.
	(ii) The quantity $-2(1-\pi_k)C$ is a class-prior–dependent \emph{constant shift} and therefore does not change the ERM minimizer; we retain it to keep \eqref{eq:emp-risk-sym} exactly unbiased for \eqref{eq:pop-risk-sym} and to enable the numerical constant--sum checks reported in our tables.
	\subsection{Class-Prior Estimation}\label{subsec:class_prior}
	In the standard contamination model for PU learning, the unlabeled marginal decomposes as
	\[
	P_X \;=\; (1-\pi)\,P_0 \;+\; \pi\,P_1,
	\]
	where $\pi:=\mathbb{P}(y=+1)$ is the class prior (positive proportion), $P_1$ is the positive class-conditional, and $P_0$ the negative one.
	Blanchard \emph{et al.}\cite{Blanchard2010SSND} connect class-prior estimation with Neyman--Pearson (NP) classification: letting $R^\star_{X,\alpha}$ denote the optimal achievable risk under a false-positive-rate constraint $R_0(f)\!\le\!\alpha$, they show that for any classifier family,
	\[
	\pi \;\ge\; 1 \;-\; \frac{R^\star_{X,\alpha}}{1-\alpha},
	\]
	and, when the class of decision rules is sufficiently rich,
	\[
	\pi \;=\; \inf_{\alpha \rightarrow 1^-}\!\left( 1 \;-\; \frac{R^\star_{X,\alpha}}{1-\alpha}\right),
	\]
	i.e., $\pi$ coincides with the slope of the ROC curve at its right endpoint. This yields a distribution-free \emph{lower bound} on $\pi$ and a principled test for ``$\pi=0$'' (novelty detection); it also clarifies that when $\pi=0$ the positive component $P_1$ is not identifiable from $P_X$ alone, motivating procedures that gracefully reduce to inductive detectors as $\pi\to 0$.
	A practical caveat is that na\"{\i}ve partial matching of $p(x)\approx \theta\,p(x\mid y{=}1)$ tends to \emph{overestimate} $\pi$ in the presence of class overlap.
	
	To address this upward bias, du Plessis \emph{et al.}~\cite{duPlessis2017ClassPriorPU} reformulate class-prior estimation via \emph{penalized $f$-divergences}, making candidate mixtures with $t>1$ infeasible and thereby removing the systematic overestimation. In particular, a penalized $L_1$ objective yields closed-form, density-estimation–free solutions together with unified bias and estimation-error bounds, and demonstrates robust, efficient estimates across diverse datasets.
	
	From an engineering standpoint, a complementary pipeline is effective: first apply the NP-based reduction to obtain a conservative \emph{lower bound} on $\pi$ (and to test for the existence of a positive component), then compute a \emph{point estimate} with the penalized $L_1$ approach to achieve numerical stability and accuracy, optionally reporting uncertainty via resampling. This combination balances identifiability and precision across varying $\pi$ and overlap regimes.
	
	\textbf{Practical pipeline (NP lower bound + penalized $L_1$ point estimate; OVR, no explicit $p_N$).}
	We provide a three-step, implementation-ready recipe aligned with our notation ($\pi_i$ for $i=1,\dots,k-1$; positives $P_i$; unlabeled mixture $U$; margin $z_i(x)=g_i(x)$). The constant--sum condition with $C=1$ allows us to eliminate explicit negatives in both theory and estimation, so all steps below use only $P_i$ and $U$.
	\begin{enumerate}
		\item \textbf{Detectability \& NP-type initial lower bounds.}
		For each observed class $i$, construct a Neyman--Pearson detector $\varphi_i:\mathcal{X}\!\to\!\{0,1\}$ that separates a positive proxy of class $i$ from the unlabeled mixture $U$. Concretely, use the margin $z_i(\boldsymbol{x})=f_i(\boldsymbol{x})$ with a threshold chosen to control the type-I error at level $\alpha$ on the proxy; let
		$\widehat{\mathbb{P}}_{U}\!\big[\varphi_i{=}1\big]$ and
		$\widehat{\mathbb{P}}_{P_i}\!\big[\varphi_i{=}1\big]$
		denote the empirical acceptance rates on $U$ and $P_i$, respectively.\footnote{If only weak positives are available, use the top-$q\%$ by margin $z_i(\boldsymbol{x})$ as the proxy.}
		Define the per-class NP lower bound
		\[
		\underline{\pi}_i \;:=\;
		\max\!\left\{\,0,\ 
		\frac{\,\widehat{\mathbb{P}}_{U}\!\big[\varphi_i{=}1\big]-\alpha\,}
		{\,\max\{\widehat{\mathbb{P}}_{P_i}\!\big[\varphi_i{=}1\big],\ \varepsilon\}\,}
		\right\},\qquad \varepsilon>0.
		\]
		If $\underline{\pi}_i=0$ across a small grid $\alpha\!\in\!\{0.01,0.02,0.05\}$, mark class $i$ as \emph{non-detectable} and fix $\pi_i\!\equiv\!0$ downstream.
		
		\item \textbf{Penalized $L_1$ point estimate via margin-moment matching (no explicit $p_N$).}
		Let $\{\psi_r\}_{r=1}^R$ be summary functions of the margin (e.g., histogram bins or smooth moments of $z_i$).
		Compute empirical moments on $U$ and $P_i$:
		\[
		\hat\mu_{U,r}=\tfrac{1}{n_U}\!\sum_{\boldsymbol{x}\in U}\psi_r\!\big(z_i(\boldsymbol{x})\big),
		\quad
		\hat\mu_{P_i,r}=\tfrac{1}{n_{P_i}}\!\sum_{\boldsymbol{x}\in P_i}\psi_r\!\big(z_i(\boldsymbol{x})\big).
		\]
		Stack $\{\hat\mu_{P_i,r}\}$ into $A\in\mathbb{R}^{R\times (k-1)}$ and $\{\hat\mu_{U,r}\}$ into $b\in\mathbb{R}^{R}$, and estimate $\pi=(\pi_1,\dots,\pi_{k-1})$ by
		\begin{align}
			\widehat{\pi}\ \in\ 
			\arg\min_{\pi\in\Delta^{k-1}}
			\ \big\|A\,\pi - b\big\|_2^2\;+\;\lambda\,\|\pi-\underline{\pi}\|_{1},
			\label{eq:pi-l1-ova}
		\end{align}
		where the feasible set
		\(
		\Delta^{k-1}
		=\big\{\pi_i\ge \underline{\pi}_i,\ \sum_{i=1}^{k-1}\pi_i\le 1\big\}
		\)
		enforces the NP lower bounds from Step~1 and the mixture constraint implied by \eqref{eq:pop-risk-sym}--\eqref{eq:emp-risk-sym}.
		We solve \eqref{eq:pi-l1-ova} using projected first-order updates; projection onto $\Delta^{k-1}$ is closed-form via per-coordinate clipping to $[\underline{\pi}_i,\infty)$ followed by renormalization to ensure $\sum_i\pi_i\le 1$.
		
		\item \textbf{Uncertainty via resampling and downstream robustness.}
		Repeat Steps~1–2 over $B$ bootstrap resamples of the calibration folds and the moment summaries to obtain $\{\widehat{\pi}^{(b)}\}_{b=1}^B$.
		Report the percentile interval at level $1-\delta$, and propagate uncertainty in two complementary ways:
		(i) fix $\widehat{\pi}$ and run a robustness sweep $\widehat{\pi}\cdot(1\pm\Delta)$ with $\Delta\in\{10\%,20\%,30\%\}$; or
		(ii) sample $\pi$ from the empirical distribution of $\{\widehat{\pi}^{(b)}\}$ during training/evaluation, and report mean$\pm$std of task metrics.
	\end{enumerate}
	\section{Theoretic Analysis}\label{sec:4}
	\subsection{Estimation Error Bound}
	We provide a high-level generalization guarantee for CSMPU. 
	Let $\mathcal{F}$ be a hypothesis class of scoring functions $f:\mathcal{X}\to\mathbb{R}^{k}$; for each $f\in\mathcal{F}$ and $y\in\mathcal{Y}$, write $f_y(\boldsymbol{x})$ for the $y$-th score assigned to input $\boldsymbol{x}$. 
	For example, $\mathcal{F}$ may consist of functions induced by neural networks $g:\mathcal{X}\to\mathbb{R}^{|\mathcal{Y}|}$ of bounded norm, with $f_y(\boldsymbol{x})=g_y(\boldsymbol{x})$.
	
	Assume the base surrogate $\ell:\mathbb{R}\times\mathcal{Y}\to\mathbb{R}$ is $L_\ell$-Lipschitz in its first argument and bounded by $C_\ell$. 
	Let $\mathcal{L}(f(\boldsymbol{x}),y)$ denote the CSMPU loss constructed from $\ell$ (see Section~\ref{sec:3}). 
	Define the population risk and its empirical counterpart by
	\[
	\mathcal{R}(f)
	\;:=\;
	\mathbb{E}_{(\boldsymbol{x},y)}\big[\mathcal{L}\!\big(f(\boldsymbol{x}),y\big)\big],
	\hat{\mathcal{R}}(f)
	\;:=\;
	\frac{1}{n}\sum_{i=1}^{n}\mathcal{L}\!\big(f(\boldsymbol{x}_i),y_i\big).
	\]
	
	For each $y\in\mathcal{Y}$, define the coordinate function class
	\[
	\mathcal{F}_y \;=\; \{\, \boldsymbol{x}\mapsto f_y(\boldsymbol{x}) \;:\; f\in\mathcal{F} \,\}.
	\]
	By standard Rademacher complexity arguments~\cite{maximov2018rademacher,musayeva2019rademacher}, we recall:
	
	\begin{definition}[Expected Rademacher complexity]
		Let $Z_1,\dots,Z_n$ be i.i.d.\ random variables drawn from a distribution $\mathcal{D}$ over a sample space $\mathcal{Z}$, and let $\mathcal{H}\subseteq\{h:\mathcal{Z}\to\mathbb{R}\}$ be a class of measurable functions. The expected Rademacher complexity of $\mathcal{H}$ is
		\[
		\Re_n(\mathcal{H})
		\;=\;
		\mathbb{E}_{Z_{1:n}\sim\mathcal{D}}
		\,\mathbb{E}_{\sigma}
		\left[
		\sup_{h\in\mathcal{H}}\;
		\frac{1}{n}\sum_{i=1}^{n}\sigma_i\,h(Z_i)
		\right],
		\]
		where $\sigma=(\sigma_1,\dots,\sigma_n)$ are i.i.d.\ Rademacher variables taking values in $\{-1,+1\}$ with equal probabilities.
	\end{definition}
	First, we have the following decomposition.
	
	\begin{lemma}
		\label{lem:excess-risk-decomp}
		The excess risk admits the bound
		\[
		\begin{split}
			\mathcal{R}(\hat f)-\mathcal{R}(f^\ast)
			\;&\le\;
			2\sum_{i=1}^{k-1}\pi_i\,\sup_{f\in\mathcal{F}}
			\bigl|\hat{\mathcal{R}}_{i}(f)-\mathcal{R}_{i}(f)\bigr| \\
			\;&+\;
			2\,\sup_{f\in\mathcal{F}}
			\bigl|\hat{\mathcal{R}}_{k}(f)-\mathcal{R}_{k}(f)\bigr| ,
		\end{split}
		\]
		where $\mathcal{R}_i(\cdot)$ and $\hat{\mathcal{R}}_i(\cdot)$ denote the population and empirical risks of the $i$-th CSMPU component, respectively.
	\end{lemma}
	
	\begin{lemma}
		\label{lem:unif-dev}
		Let $C_{\mathcal{L}}:=\sup_{\boldsymbol{x}\in\mathcal{X},\,f\in\mathcal{F},\,y\in\mathcal{Y}}
		\mathcal{L}\big(f(\boldsymbol{x}),y\big)$.
		Then, for any $i\in\{1,\dots,k-1\}$ and any $\delta>0$, with probability at least $1-\delta$,
		\[
		\sup_{f\in\mathcal{F}}
		\bigl|\hat{\mathcal{R}}_{i}(f)-\mathcal{R}_{i}(f)\bigr|
		\;\le\;
		2\,\bar{\Re}_{n_i}\!\big(\mathcal{H}_i\big)
		\;+\;
		2\,C_{\mathcal{L}}\sqrt{\frac{2\log(2/\delta)}{n_i}},
		\]
		where $\bar{\Re}_{n_i}(\mathcal{H}_i)$ is the (empirical) Rademacher complexity of the coordinate class
		$\mathcal{H}_i=\{\,\boldsymbol{x}\mapsto f_i(\boldsymbol{x}) : f\in\mathcal{F}\,\}$
		under $p_i$ and sample size $n_i$.
	\end{lemma}
	
	By Lemma~\ref{lem:excess-risk-decomp}, the excess risk decomposes class-wise, reducing the analysis to uniform deviations between empirical and population risks for each CSMPU component.
	Using the $L_\ell$-Lipschitz and boundedness assumptions on the base surrogate $\ell$, Lemma~\ref{lem:unif-dev} together with standard symmetrization and contraction yields, for each labeled component $i\in\{1,\dots,k-1\}$,
	\[
	\bar{\mathfrak{R}}_{n_i,p_i}(\mathcal{H}_i)
	\;\le\;
	2L_\ell\!\left(
	\mathfrak{R}_{n_i,p_i}(\mathcal{G}_i)
	\;+\;
	\mathfrak{R}_{n_i,p_i}(\mathcal{G}_k)
	\right),
	\]
	accompanied by Hoeffding-type concentration terms controlled by $C_\ell$.
	Substituting these bounds into Lemma~\ref{lem:excess-risk-decomp}, applying a union bound over the $(k\!-\!1)$ positive components and the unlabeled component, and collecting constants yields Theorem~\ref{thm:csmpu_generalization}.
	\begin{theorem}
		\label{thm:csmpu_generalization}
		With probability at least $1-\delta$ over the sample draw,
		\begin{equation*}
			\begin{split}
				&\mathcal{R}(\hat f)-\mathcal{R}(f^\ast)
				\;\\
				& \le\! 
				4\!\sum_{i=1}^{k-1}\!\pi_i \!\left(
				2L_{\ell}\!\big(\Re_{n_i,p_i}\!(\mathcal{G}_i)\!+\!\Re_{n_i,p_i}(\mathcal{G}_k)\big)
				\!+\! C_{\mathcal{L}}\!\sqrt{\frac{2\log(2/\delta)}{\,n_i\,}}
				\right)
				\\
				&\quad+\;
				4\left(
				L_{\ell}\big(\Re_{n_k,p_u}(\mathcal{G}_i)\!+\!\Re_{n_k,p_u}(\mathcal{G}_k)\big)
				\!+\! C_{\mathcal{L}}\sqrt{\frac{\log(2/\delta)}{\,n_k\,}}
				\right),
			\end{split}
		\end{equation*}
	\end{theorem}
	
	\noindent
	Here, $\hat f\in\arg\min_{f\in\mathcal{F}}\hat{\mathcal{R}}(f)$ is the empirical minimizer and $f^\ast\in\arg\min_{f\in\mathcal{F}}\mathcal{R}(f)$ is the population minimizer; $\Re_{n_i,p_i}(\cdot)$ is the Rademacher complexity w.r.t.\ class-$i$ distribution $p_i$ and sample size $n_i$; $n_k$ is the unlabeled sample size and $\Re_{n_k,p_u}(\cdot)$ is computed w.r.t.\ the unlabeled marginal $p_u$. A detailed proof is provided in the Appendix III.
	
	\paragraph*{Discussion}
	The bound implies \emph{consistency}: as $n_i,n_k\to\infty$, we have $\mathcal{R}(\hat f)\to \mathcal{R}(f^\ast)$ with high probability, under the stated capacity assumptions. It also highlights the dependencies on the function-class complexity and per-class data: tighter complexities (smaller Rademacher terms) and larger $n_i,n_k$ yield smaller excess risk. In Section~\ref{sec:4} we empirically verify that CSMPU generalizes well and avoids overfitting in line with this theory.
	
	\subsection{Risk bias under class-prior misspecification}
	Let $\mathcal{O}\subseteq\{1,\dots,k-1\}$ denote the set of observed classes. 
	For each $i\in\mathcal{O}$, the per-class OVR population risk admits a PU-style linear form in the class prior $\pi_i$:
	\[
	\begin{split}
		\mathcal{R}_i(f)
		\;&=\;
		\underbrace{\mathbb{E}_{\boldsymbol{x}\sim p_u}\!\big[\ell_i^{U}\!\big(f_i(\boldsymbol{x})\big)\big]}_{\text{unlabeled term}} \\
		\;&\qquad +\;\pi_i\,
		\underbrace{\mathbb{E}_{\boldsymbol{x}\sim p_i}\!\big[\Delta\ell_i\!\big(f_i(\boldsymbol{x})\big)\big]}_{\,b_i(f)} 
		\;+\;\text{const},
	\end{split}
	\]
	where $p_i(\boldsymbol{x})=p(\boldsymbol{x}\mid y=i)$ and $p_u$ is the unlabeled marginal; 
	$\Delta\ell_i(z):=\ell_i^{+}(z)-\ell_i^{-}(z)$ is the class-$i$ loss gap (e.g., induced by a constant–sum base loss), and $b_i(f)$ abbreviates the corresponding expectation.
	Let $\widehat{\pi}_i$ be the (possibly misspecified) prior used in the estimator, and let $\lambda_i\ge 0$ be the cost weights from the cost-sensitive OVR objective.
	The class-prior misspecification bias for class $i$ is
	\[
	\mathrm{Bias}_i(f;\widehat{\pi}_i,\pi_i)
	\;:=\;
	\big|\mathcal{R}_i(f)-\mathcal{R}_i^{(\widehat{\pi}_i)}(f)\big|
	\;=\;
	\big|\,\widehat{\pi}_i-\pi_i\,\big|\cdot\big|\,b_i(f)\,\big|.
	\]
	
	\begin{lemma}[Uniform bound on misspecification bias]\label{lem:prior-bias}
		Assume the base loss is bounded, i.e., $\ell_i^{+}(z),\ell_i^{-}(z)\in[0,C_\ell]$ for all $z$ and $i$.
		Then for any $f$,
		\[
		\begin{split}
			&\mathrm{Bias}_i(f;\widehat{\pi}_i,\pi_i)
			\;\le\;
			C_{\Delta}\,\big|\,\widehat{\pi}_i-\pi_i\,\big|,
			\\
			&\qquad
			C_{\Delta}
			:= \sup_{f}\big|\mathbb{E}_{\boldsymbol{x}\sim p_i}\!\big[\Delta\ell_i\!\big(f_i(\boldsymbol{x})\big)\big]\big|
			\;\le\; 2C_\ell.
		\end{split}
		\]
		Consequently, the total (cost-weighted) population bias satisfies
		\[
		\sum_{i\in\mathcal{O}}\lambda_i\,\mathrm{Bias}_i(f;\widehat{\pi}_i,\pi_i)
		\;\le\;
		\Big(\sum_{i\in\mathcal{O}}\lambda_i\Big)\,C_{\Delta}\;
		\max_{i\in\mathcal{O}}\big|\,\widehat{\pi}_i-\pi_i\,\big|.
		\]
		If an outer non-negative correction $\varphi$ with Lipschitz constant $L_\varphi$ (e.g., $\mathrm{ReLU}$ or $\mathrm{ABS}$, both with $L_\varphi=1$) is applied to the aggregated risk, the same bounds hold multiplied by $L_\varphi$.
	\end{lemma}
	
	\noindent [Proof sketch]
		By linearity in $\pi_i$,
		$\big|\mathcal{R}_i(f)-\mathcal{R}_i^{(\widehat{\pi}_i)}(f)\big|
		=\big|\,\widehat{\pi}_i-\pi_i\,\big|\cdot\big|\mathbb{E}_{\boldsymbol{x}\sim p_i}[\Delta\ell_i(f_i(\boldsymbol{x}))]\big|$.
		Since $|\Delta\ell_i(z)|\le |\ell_i^{+}(z)|+|\ell_i^{-}(z)|\le 2C_\ell$, we have $C_\Delta\le 2C_\ell$.
		A $L_\varphi$-Lipschitz outer map scales these inequalities by at most $L_\varphi$.
	
	\begin{theorem}[Excess risk with class-prior misspecification]\label{thm:excess-misspec}
		Let $\widehat{f}$ be an ERM for the empirical corrected objective constructed with priors $\widehat{\pi}_i$ and sample sizes as in Section~\ref{sec:4}. 
		With probability at least $1-\delta$,
		\[
		\begin{split}
			\mathcal{R}(\widehat{f})-&\mathcal{R}(f^\star)
			 \;\le\;
			\underbrace{2\,\Gamma_{n,m}(\delta)}_{\text{uniform deviation (Rademacher) term}} \\
			& \;+\;
			\underbrace{\sum_{i\in\mathcal{O}}\lambda_i\!\Big(\mathrm{Bias}_i(\widehat{f};\widehat{\pi}_i,\pi_i)
				+\mathrm{Bias}_i(f^\star;\widehat{\pi}_i,\pi_i)\Big)}_{\text{prior misspecification bias}}.
		\end{split}
		\]
		In particular, under Lemma~\ref{lem:prior-bias} and $L_\varphi\le 1$,
		\[
		\mathcal{R}(\widehat{f})-\mathcal{R}(f^\star)
		\;\le\;
		2\,\Gamma_{n,m}(\delta)
		\;+\;
		2\,C_{\Delta}\sum_{i\in\mathcal{O}}\lambda_i\,\big|\,\widehat{\pi}_i-\pi_i\,\big|.
		\]
	\end{theorem}
	The completely proof is provided in Appendix IV.
	\begin{remark}[Single-number summary]
		If a scalar tolerance $\varepsilon:=\max_{i\in\mathcal{O}}\big|\,\widehat{\pi}_i-\pi_i\,\big|$ is available and $\sum_{i\in\mathcal{O}}\lambda_i=1$, Theorem~\ref{thm:excess-misspec} yields
		\[
		\mathcal{R}(\widehat{f})-\mathcal{R}(f^\star)
		\;\le\;
		2\,\Gamma_{n,m}(\delta)
		\;+\;
		2\,C_{\Delta}\,\varepsilon,
		\]
		i.e., the excess risk degrades at most \emph{linearly} in the worst-case prior error, with slope controlled by the loss range.
	\end{remark}
	
	\section{The Corrected Function}
	\label{sec:corrected}
	\subsection{Motivation and Construction}
	In finite samples, the unbiased empirical risk $\hat{\mathcal{R}}(f)$ from Section~\ref{sec:3} can become negative due to the constant offset $-2(1-\pi_k)$ and sampling fluctuations, even though the population risk is nonnegative. This may destabilize optimization and encourage overfitting. To mitigate this, we \emph{post-compose} the outer (classwise-aggregated) objective with a hard nonnegativity correction using the standard $\mathrm{ReLU}$,
	\[
	\mathrm{ReLU}(z)\;=\;\max\{z,0\}.
	\]
	Following MPU practice, the correction is applied at the \emph{outer} risk level—exactly where negativity can arise—while leaving the inner per-sample losses unchanged.
	
	Formally, for $i=1,\dots,k-1$ define
	\[
	\mu_i\;=\;\mathbb{E}_{\boldsymbol{x}\sim p_i}\!\big[\mathcal{L}_i(f(\boldsymbol{x}))\big],
	\qquad
	\hat\mu_i\;=\;\frac{1}{n_i}\sum_{j=1}^{n_i}\mathcal{L}_i\!\big(f(\boldsymbol{x}_{i,j})\big),
	\]
	and for the unlabeled mixture $p_u$,
	\[
	\mu_u\;=\;\mathbb{E}_{\boldsymbol{x}\sim p_u}\!\big[\mathcal{L}_k(f(\boldsymbol{x}))\big],
	\qquad
	\hat\mu_u\;=\;\frac{1}{n_k}\sum_{j=1}^{n_k}\mathcal{L}_k\!\big(f(\boldsymbol{x}_{k,j})\big),
	\]
	Let $g(z)=\mathrm{ReLU}(z)$ and consider
	\[
	\begin{split}
		\mathcal{R}(f)\;&=\;\sum_{i=1}^{k-1}2\pi_i\,g(\mu_i)\;+\;g\!\big(\mu_u-2(1-\pi_k)\big), \\
		\hat{\mathcal{R}}(f)\;&=\;\sum_{i=1}^{k-1}2\pi_i\,g(\hat\mu_i)\;+\;g\!\big(\hat\mu_u-2(1-\pi_k)\big).
	\end{split}
	\]
	The \emph{corrected empirical risk} is then
	\[
	\widetilde{\mathcal{R}}(f)\;:=\;\mathrm{ReLU}\!\big(\hat{\mathcal{R}}(f)\big).
	\]
	Because $\mathrm{ReLU}$ is convex and $1$-Lipschitz, this outer correction preserves the bias and concentration properties used in our analysis. Throughout we assume: (i) samples across $\{\mathcal{X}_i\}_{i=1}^{k-1}$ and $\mathcal{X}_k$ are independent; (ii) the base loss is bounded, $\ell\in[0,C_\ell]$; and (iii) $g=\mathrm{ReLU}$ (convex, $1$-Lipschitz).
	\subsection{Concentration for the Downward Deviation}
	We relate the corrected objective $\widetilde{\mathcal{R}}:=g(\hat{\mathcal{R}})$ to the population risk $\mathcal{R}$ for a convex $1$-Lipschitz map $g$.
	Define the downward–deviation event after correction
	\[
	\Omega^-(g)\;:=\;\{\,g(\hat{\mathcal{R}})-g(\mathcal{R})<0\,\},
	\qquad
	\Omega^{-c}(g):=\big(\Omega^-(g)\big)^{c},
	\]
	where the superscript $c$ denotes the \emph{complement}. On $\Omega^{-c}(g)$ no downward deviation occurs.
	
	\begin{theorem}[corrected risk with convex $1$-Lipschitz $g$]
		\label{thm:gen-corrected}
		Under the assumptions above and $\ell\in[0,C_\ell]$,
		\[\mathbb{E}\big[\widetilde{\mathcal{R}}(f)-\mathcal{R}(f)\big]
		\ \le\
		B_{\mathrm{sup}}\cdot \Pr\!\big(\Omega^-(g)\big),\]
		where
		\[\begin{split}
			B_{\mathrm{sup}}
			\;& :=\;
			\sup_{(\mathcal D_L,\mathcal D_U)\in \Omega^-(g)}
			\Big(\,|\widehat{\mathcal{R}}(f)|+\big|\,\widehat{\mathcal{R}}(f)-\mathcal{R}(f)\,\big)\Big) \\
			\ &\le\
			2C_\ell \sum_{i=1}^{k-1}\pi_i\;+\;C_\ell\;+\;2(1-\pi_k).
		\end{split}\]
		Consequently, for any $\varepsilon\le \mathbb{E}\!\left[\widehat{\mathcal{R}}(f)-\mathcal{R}(f)\right]$,
		\[\begin{split}
			& \mathbb{E}\big[\widetilde{\mathcal{R}}(f)-\mathcal{R}(f)\big]
			\ \le\ \\
			& \Big(2C_\ell(3-2\pi_k)+2(1-\pi_k)\Big)\,
			\exp\!\left(
			-\frac{2\varepsilon^2}{\,C_\ell^{\,2}\Big(4\sum_{i=1}^{k-1}\tfrac{\pi_i^2}{n_i}+\tfrac{1}{n_k}\Big)}
			\right).
		\end{split}\]
		Moreover, for any $\delta\in(0,1)$, with probability at least $1-\delta$,
		\[
		\mathcal{R}(f)
		\ \le\
		\widetilde{\mathcal{R}}(f)
		\;+\;
		C_\ell\sqrt{\tfrac{1}{2}\Big(4\sum_{i=1}^{k-1}\tfrac{\pi_i^2}{n_i}+\tfrac{1}{n_k}\Big)\log\tfrac{1}{\delta}}\,.
		\]
	\end{theorem}
	The completely proof is provided in Appendix V.
	\begin{figure*}[!htbp]
		\centering
		\scriptsize
		\begin{tabular}{ccc}
			\includegraphics[width=5.7cm]{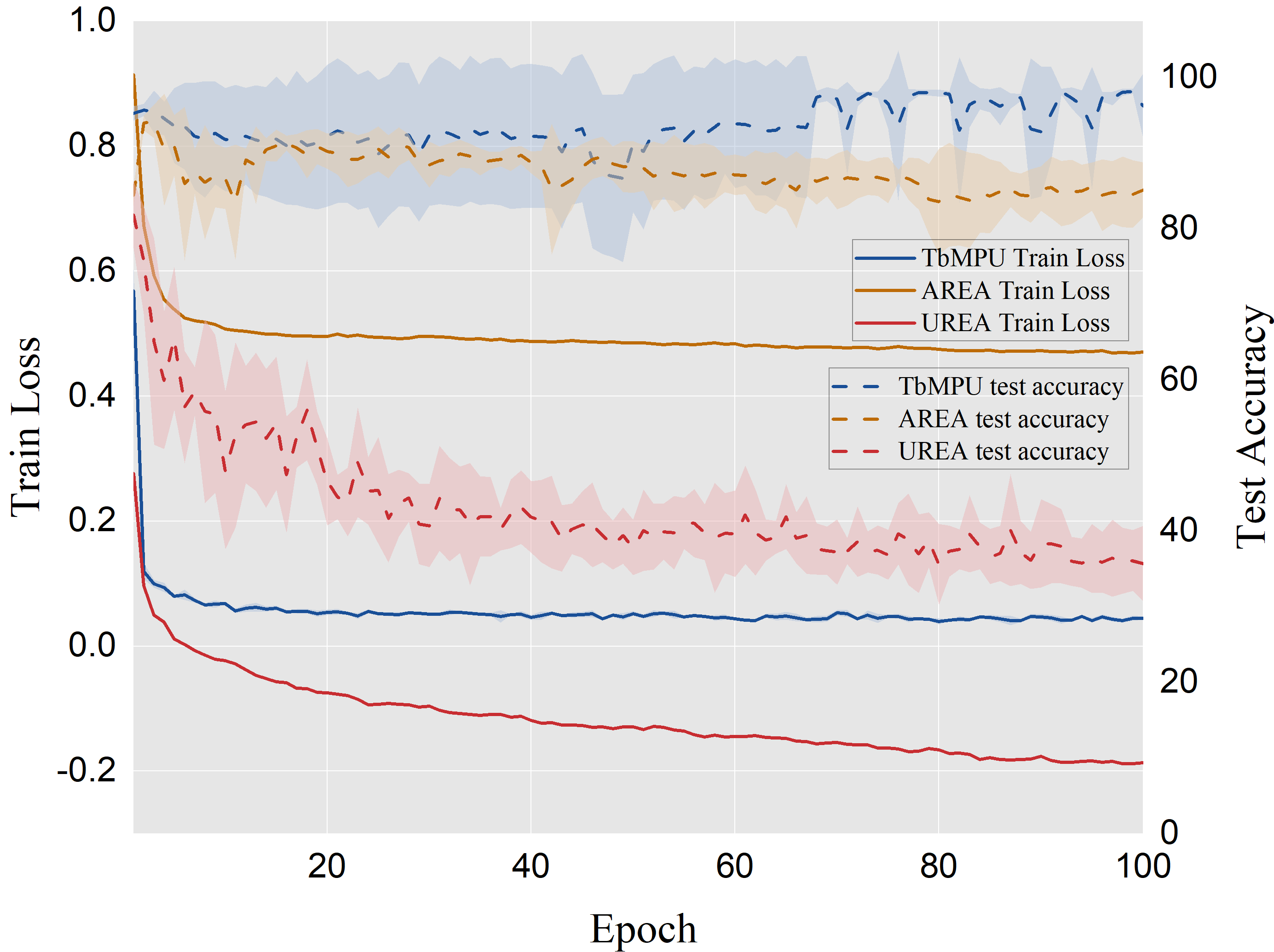} & \includegraphics[width=5.7cm]{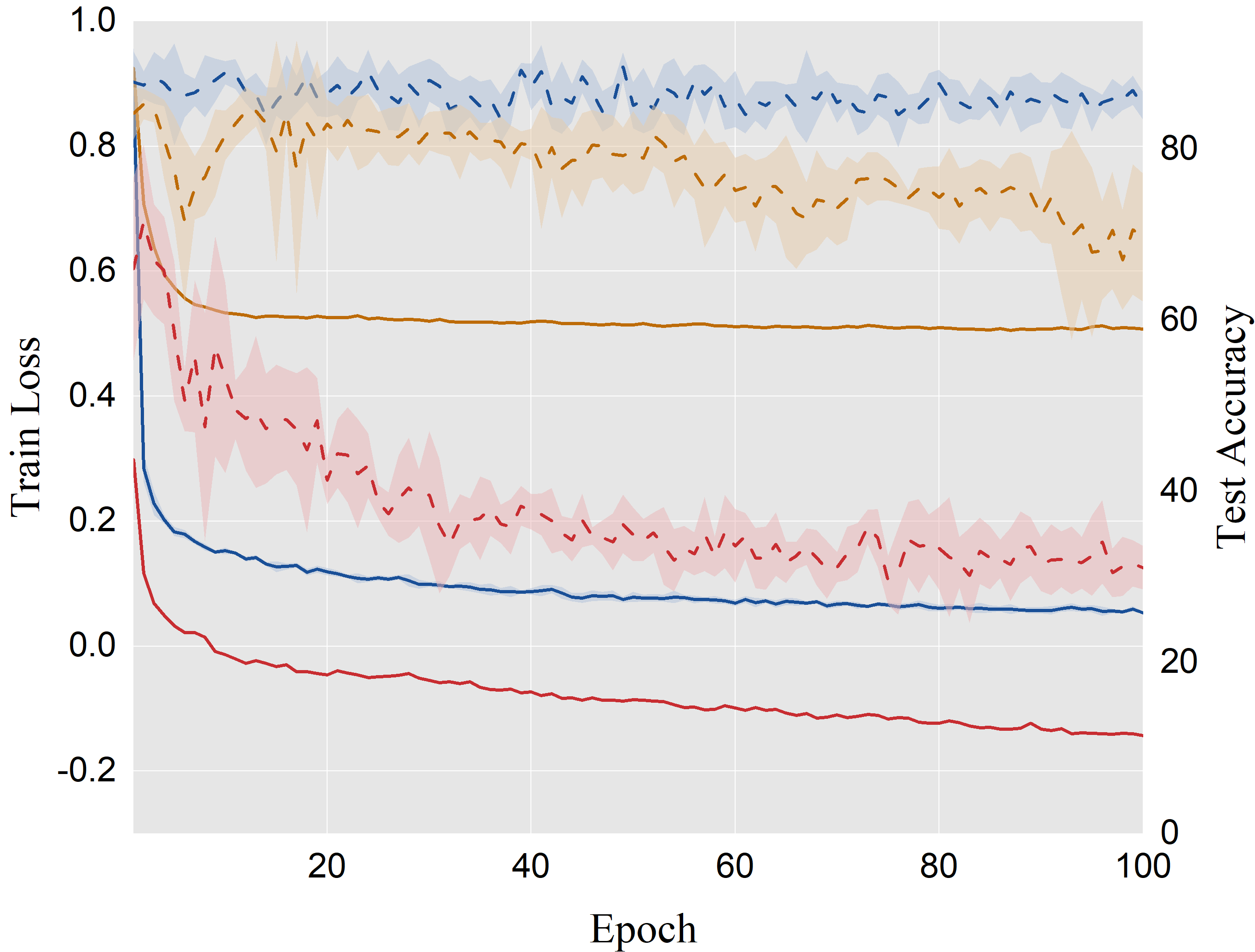} &
			\includegraphics[width=5.7cm]{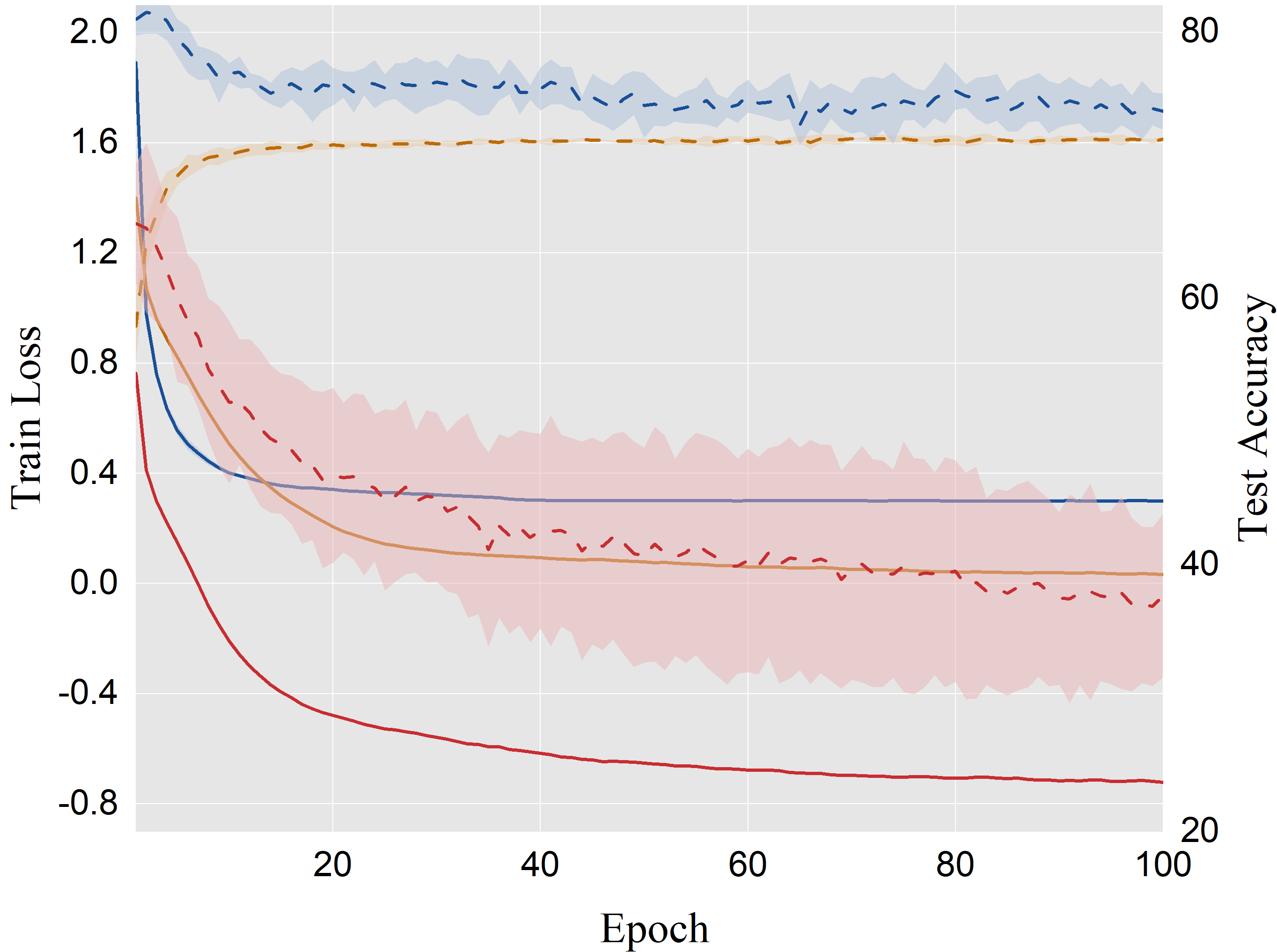} \\
			MNIST,MLP  & Fashion-MNIST,MLP & KMNIST,MLP \\
		\end{tabular}
		\caption{Training and test curves of CSMPU on three datasets (MNIST, FashionMNIST, and KMNIST) with negative-class prior $\pi_k=0.2$. For each dataset, we plot training loss, test loss, training accuracy, and test accuracy versus epochs; curves are averaged over five independent runs. CSMPU exhibits smooth convergence without pronounced overfitting, contrasting with representative MPU baselines.}
		\label{fig:compared}
	\end{figure*}
	\begin{table*}[h!]
		\begin{center}
			\caption{Classification accuracy (mean $\pm$ standard deviation) across benchmark datasets with negative-class prior $\pi_k=0.2$. The best result per dataset is shown in \textbf{bold}. Each entry averages five independent runs. \textit{ABS} uses an absolute-value correction as the outer function, whereas \textit{NN} applies a ReLU-based nonnegativity correction.}
			\label{results}
			\setlength{\tabcolsep}{7pt}
			\fontsize{9}{10}\selectfont
			\renewcommand{\arraystretch}{1}
			\begin{tabular}{llllllll}
				\hline
				\multirow{2}{*}{dataset} & \multirow{2}{*}{N.} & \multicolumn{3}{l}{proposed method}&\multicolumn{3}{l}{baseline} \\
				&&ABS&NN&\URE&AREA&biased super&\URE \\
				\hline
				\multirow{3}{*}{Fashion-MNIST}&4&\textbf{89.82 $\pm$ 0.33}&85.09 $\pm$ 1.44&72.61  $\pm$  0.26&69.71  $\pm$  6.69&31.10  $\pm$  2.27&51.92  $\pm$  1.72\\
				&6&\textbf{81.28 $\pm$ 1.34}&76.36 $\pm$ 1.33&29.24 $\pm$ 1.36&52.45  $\pm$  3.68&24.55  $\pm$  4.50&38.68  $\pm$  1.91\\
				&8&\textbf{77.89 $\pm$ 0.16}&76.38 $\pm$ 1.11&47.60 $\pm$ 2.85&51.35  $\pm$  2.76&20.32  $\pm$  1.09&26.80  $\pm$  2.07\\
				\hline
				\multirow{3}{*}{KMNIST}&4&\textbf{87.59 $\pm$ 0.25}&86.26 $\pm$ 0.29&74.34 $\pm$ 0.78&74.03  $\pm$  1.22&37.68  $\pm$  5.49&77.04  $\pm$  1.30\\
				&6&\textbf{71.69 $\pm$ 0.79}&61.32 $\pm$ 2.86&24.29 $\pm$ 0.12&64.14  $\pm$  1.24&32.90  $\pm$  1.31&22.77  $\pm$  0.51\\
				&8&\textbf{78.19 $\pm$ 0.29}&71.13 $\pm$ 0.35&38.85 $\pm$ 0.30&69.13  $\pm$  2.79&41.03  $\pm$  0.78&39.55  $\pm$  0.61\\
				\hline
				\multirow{3}{*}{MNIST}&4&\textbf{95.02 $\pm$ 1.05}&94.53 $\pm$ 0.79&89.33 $\pm$ 0.46&85.18  $\pm$  3.25&35.75  $\pm$  4.44&95.08  $\pm$  0.95\\
				&6&\textbf{89.80 $\pm$ 0.76}&86.88 $\pm$ 0.70&36.35 $\pm$ 0.81&79.03  $\pm$  2.51&23.61  $\pm$  2.27&92.75  $\pm$  1.06\\
				&8&\textbf{91.42 $\pm$ 1.10}&90.47 $\pm$ 1.00&46.24 $\pm$ 0.93&73.75  $\pm$  3.05&20.14  $\pm$  2.35&94.57  $\pm$  0.47\\
				\hline
				
				\multirow{3}{*}{PBRHD}&4&\textbf{97.33 $\pm$ 1.11}&96.84 $\pm$ 0.56&89.76 $\pm$ 1.46&91.90  $\pm$  1.34&76.68  $\pm$  2.99&84.81  $\pm$  0.37\\
				&6&\textbf{88.60 $\pm$ 0.76}&85.89 $\pm$ 2.92&37.72 $\pm$ 1.97&82.66  $\pm$  1.84&68.17  $\pm$  1.42&83.33  $\pm$  0.73\\
				&8&\textbf{90.40 $\pm$ 1.54}&90.29 $\pm$ 1.09&53.69 $\pm$ 2.15&84.61  $\pm$  1.07&74.87  $\pm$  1.56&85.49  $\pm$  0.76\\
				\hline
				\multirow{3}{*}{Semeion}&4&90.18 $\pm$ 2.03&\textbf{90.96 $\pm$ 2.03}&86.56 $\pm$ 2.85&70.62  $\pm$  3.08&76.11  $\pm$  4.57&74.01  $\pm$  4.25\\
				&6&\textbf{70.44 $\pm$ 1.75}&63.23 $\pm$ 4.25&44.67 $\pm$ 3.18&73.49  $\pm$  2.27&60.05  $\pm$  4.45&70.48  $\pm$  3.23\\
				&8&75.67 $\pm$ 0.55&\textbf{75.93 $\pm$ 0.65}&62.03 $\pm$ 3.00&74.00  $\pm$  2.39&63.88  $\pm$  1.93&73.79  $\pm$  1.80\\
				\hline
				\multirow{3}{*}{SVHN}&4&\textbf{79.08 $\pm$ 1.77}&57.09  $\pm$  16.06&66.65  $\pm$  7.23&74.04  $\pm$  6.02&26.97  $\pm$  2.03&67.92  $\pm$  11.09\\
				&6&\textbf{83.50  $\pm$  2.86}&63.35  $\pm$  10.11&51.61  $\pm$  20.35&68.29  $\pm$  2.43&63.92  $\pm$  0.84&34.26  $\pm$  2.82\\
				&8&\textbf{87.83  $\pm$  1.40}&52.73  $\pm$  25.71&42.16  $\pm$  21.15&64.71  $\pm$  3.24&11.26  $\pm$  2.38&58.25  $\pm$  3.44\\
				\hline
				\multirow{3}{*}{USPS}&4&\textbf{94.90 $\pm$ 0.13}&94.22 $\pm$ 0.38&87.37 $\pm$ 1.03&92.22  $\pm$  0.69&48.77  $\pm$  4.23&92.32  $\pm$  1.28\\
				&6&\textbf{84.81 $\pm$ 1.13}&81.12 $\pm$ 1.95&48.20 $\pm$ 0.58&79.50  $\pm$  2.38&62.27  $\pm$  3.44&89.03  $\pm$  0.89\\
				&8&\textbf{85.84 $\pm$ 0.31}&83.55 $\pm$ 0.28&48.70 $\pm$ 1.31&81.36  $\pm$  1.66&64.41  $\pm$  2.56&89.06  $\pm$  1.68\\
				\hline
				{WaveForm1}&3&82.45 $\pm$ 0.33&\textbf{84.45 $\pm$ 1.23}&82.79 $\pm$ 1.05&75.00  $\pm$  4.52&24.38  $\pm$  2.99&75.84  $\pm$  6.48\\
				\hline
			\end{tabular}
		\end{center}
	\end{table*}
	\begin{table*}[h!]
		\begin{center}
			\caption{Classification accuracy (mean ± standard deviation) across benchmark datasets with negative class prior $\pi_k=0.5$. The best result per dataset is highlighted in bold. Each result is averaged over five independent runs.}
			\label{results1}
			\setlength{\tabcolsep}{7pt}
			\fontsize{9}{10}\selectfont
			\renewcommand{\arraystretch}{1}
			\begin{tabular}{llllllll}
				\hline
				\multirow{2}{*}{dataset} & \multirow{2}{*}{N.} & \multicolumn{3}{l}{proposed method}&\multicolumn{3}{l}{baseline}\\
				&&ABS&NN&\URE&AREA&biased super&\URE\\
				\hline
				\multirow{3}{*}{Fashion-MNIST}&4& \textbf{90.59  $\pm$  1.14}	& 89.62  $\pm$  0.75	& 87.38  $\pm$  0.34&80.90  $\pm$  2.15&39.13  $\pm$  7.77&50.30  $\pm$  1.23\\
				&6& 86.58  $\pm$  0.78	& \textbf{87.95  $\pm$  0.31}	& 87.28  $\pm$  1.21&76.43  $\pm$  4.37&29.88  $\pm$  2.03&43.12  $\pm$  4.28\\
				&8& 79.53  $\pm$  1.53	& \textbf{80.19  $\pm$  0.96}	& 78.76  $\pm$  3.60	&65.39  $\pm$  2.21&20.11  $\pm$  2.42&31.39  $\pm$  5.71\\
				\hline
				\multirow{3}{*}{KMNIST}&4& \textbf{88.43  $\pm$  0.15}	& 87.01  $\pm$  0.99	& 70.93  $\pm$  1.10	&72.85  $\pm$  1.38&32.41  $\pm$  3.11&71.56  $\pm$  1.85\\
				&6& \textbf{84.44  $\pm$  0.65}	& 82.55  $\pm$  0.36	& 72.46  $\pm$  0.28	&66.20  $\pm$  0.92&53.46  $\pm$  2.89&78.46  $\pm$  1.18\\
				&8& \textbf{81.65  $\pm$  0.11}	& 78.86  $\pm$  0.43	& 72.23  $\pm$  0.39&54.39  $\pm$  2.07&58.77  $\pm$  0.83&81.09  $\pm$  0.34\\
				\hline
				\multirow{3}{*}{MNIST}&4& \textbf{93.59  $\pm$  0.57}	& 93.10  $\pm$  0.20	& 86.68  $\pm$  1.43&93.72  $\pm$  1.74&36.13  $\pm$  5.17&95.82  $\pm$  0.53\\
				&6& \textbf{97.64  $\pm$  0.17}	& 97.37  $\pm$  0.20	& 95.98  $\pm$  0.28&93.75  $\pm$  0.26&23.85  $\pm$  2.02&86.37  $\pm$  1.25\\
				&8& \textbf{97.35  $\pm$  0.43}	& 97.23  $\pm$  0.04	& 96.58  $\pm$  0.07&76.15  $\pm$  1.65&21.93  $\pm$  4.03&86.30  $\pm$  2.55\\
				\hline
				\multirow{3}{*}{PBRHD}&4& 94.82  $\pm$  0.28	& \textbf{95.70  $\pm$  0.73}	& 83.93  $\pm$  1.20&95.60  $\pm$  0.63&81.93  $\pm$  4.33&91.12  $\pm$  0.85\\
				&6& \textbf{97.63  $\pm$  0.10}	& 97.52  $\pm$  0.40	& 95.95  $\pm$  1.01	&95.78  $\pm$  0.73&87.18  $\pm$  3.46&92.02  $\pm$  0.96\\
				&8& \textbf{97.30  $\pm$  0.54}	& 96.89  $\pm$  0.79	& 93.54  $\pm$  1.04	&95.54  $\pm$  0.71&85.83  $\pm$  2.35&91.88  $\pm$  1.27\\
				\hline
				\multirow{3}{*}{Semeion}&4& 89.41  $\pm$  2.85	& \textbf{92.25  $\pm$  0.63}	& 87.86  $\pm$  1.32&85.25  $\pm$  2.46&83.07  $\pm$  2.99&88.44  $\pm$  2.26\\
				&6& \textbf{88.14  $\pm$  1.26}	& 83.85  $\pm$  2.67	& 85.22  $\pm$  1.47	&81.92  $\pm$  1.76&73.02  $\pm$  5.82&84.31  $\pm$  1.79\\
				&8& 80.95  $\pm$  3.34	& \textbf{81.85  $\pm$  0.55}	& 79.79  $\pm$  0.65	&76.16  $\pm$  1.81&61.81  $\pm$  3.40&80.58  $\pm$  0.92\\
				\hline
				\multirow{3}{*}{SVHN}&4&\textbf{87.17  $\pm$  1.30}&85.23  $\pm$  13.50&61.31  $\pm$  16.55&85.31  $\pm$  7.08&21.58  $\pm$  1.27&71.97  $\pm$  0.95\\
				&6&\textbf{88.71  $\pm$  1.96}&78.48  $\pm$  5.40&40.97  $\pm$  9.54&85.16  $\pm$  0.92&13.54  $\pm$  0.69&82.32  $\pm$  0.88\\
				&8&\textbf{87.81  $\pm$  1.13}&80.47  $\pm$  26.66&32.05  $\pm$  13.22&80.66  $\pm$  1.89&8.81  $\pm$  0.01&85.40  $\pm$  0.33\\
				\hline
				\multirow{3}{*}{USPS}&4& 92.30  $\pm$  1.22	& 93.01  $\pm$  0.52	& 92.88  $\pm$  0.79	&\textbf{93.11  $\pm$  2.00}&57.59  $\pm$  9.52&86.34  $\pm$  1.74\\
				&6& 93.12  $\pm$  0.29	& 92.90  $\pm$  0.13	& 86.09  $\pm$  0.62	&\textbf{93.23  $\pm$  0.63}&69.70  $\pm$  2.62&89.00  $\pm$  1.76\\
				&8& \textbf{90.57  $\pm$  0.26}	& 90.06  $\pm$  0.43	& 83.33  $\pm$  0.86	&90.50  $\pm$  0.19&82.20  $\pm$  2.57&88.22  $\pm$  1.47\\
				\hline
				{WaveForm1}&3& \textbf{83.59  $\pm$  0.91}	& 82.29  $\pm$  1.32	& 81.12  $\pm$  1.06	&76.30  $\pm$  3.91&24.11  $\pm$  2.15&64.49  $\pm$  2.27\\
				\hline
			\end{tabular}
		\end{center}
	\end{table*}
	\begin{table*}[h!]
		\begin{center}
			\caption{Classification accuracy (mean ± standard deviation) across benchmark datasets with negative class prior $\pi_k=0.8$. The best result per dataset is highlighted in bold. Each result is averaged over five independent runs.}
			\label{results2}
			\setlength{\tabcolsep}{7pt}
			\fontsize{9}{10}\selectfont
			\renewcommand{\arraystretch}{1}
			\begin{tabular}{llllllll}
				\hline
				\multirow{2}{*}{dataset} & \multirow{2}{*}{N.} & \multicolumn{3}{l}{proposed method}&\multicolumn{3}{l}{baseline} \\
				&&ABS&NN&\URE&AREA&biased super&\URE \\
				\hline
				\multirow{3}{*}{Fashion-MNIST}&4& 94.03  $\pm$  0.08		& \textbf{94.19  $\pm$  0.35}		& 93.72  $\pm$  0.33	&84.38  $\pm$  2.56&34.23  $\pm$  9.72&46.30  $\pm$  9.27\\
				&6& \textbf{81.80  $\pm$  3.02}		& 80.77  $\pm$  1.86		& 79.09  $\pm$  0.90	&74.65  $\pm$  5.83&25.00  $\pm$  2.05&44.04  $\pm$  4.96\\
				&8& \textbf{68.34  $\pm$  3.88}		& 67.58  $\pm$  5.60		& 68.11  $\pm$  1.86	&69.93  $\pm$  1.50&20.73  $\pm$  6.49&33.01  $\pm$  4.11\\
				\hline
				\multirow{3}{*}{KMNIST}&4& \textbf{91.71  $\pm$  0.55}		& 90.18  $\pm$  0.64		& 88.16  $\pm$  0.65	&82.73  $\pm$  4.08&69.35  $\pm$  4.49&91.71  $\pm$  0.40\\
				&6& \textbf{83.54  $\pm$  0.99}		& 82.04  $\pm$  0.60		& 79.83  $\pm$  0.23	&75.31  $\pm$  1.26&48.83  $\pm$  7.70&90.24  $\pm$  0.43\\
				&8& \textbf{78.02  $\pm$  1.30}		& 76.69  $\pm$  0.44		& 74.59  $\pm$  0.90	&62.36  $\pm$  1.96&48.68  $\pm$  7.02&89.10  $\pm$  0.86\\
				\hline
				\multirow{3}{*}{MNIST}&4& \textbf{98.61  $\pm$  0.07}		& 98.32  $\pm$  0.16		& 98.04  $\pm$  0.21	&91.95  $\pm$  1.19&28.24  $\pm$  1.51&90.61  $\pm$  1.04\\
				&6& \textbf{97.69  $\pm$  0.51}		& 97.54  $\pm$  0.28		& 96.72  $\pm$  0.64	&93.10  $\pm$  0.55&26.38  $\pm$  5.87&87.80  $\pm$  1.79\\
				&8& \textbf{96.73  $\pm$  0.41}		& 96.17  $\pm$  0.37		& 95.74  $\pm$  0.27	&84.29  $\pm$  1.56&21.41  $\pm$  2.01&84.23  $\pm$  0.33\\
				\hline
				\multirow{3}{*}{PBRHD}&4& \textbf{98.48  $\pm$  0.34}		& 97.39  $\pm$  0.67		& 98.31  $\pm$  0.33	&96.95  $\pm$  0.54&92.70  $\pm$  1.81&93.10  $\pm$  0.55\\
				&6& \textbf{97.44  $\pm$  0.19}		& 97.04  $\pm$  1.00		& 97.44  $\pm$  0.26	&97.62  $\pm$  0.28&92.37  $\pm$  0.89&93.78  $\pm$  0.87\\
				&8& \textbf{96.84  $\pm$  0.14}		& 96.70  $\pm$  0.35		& 96.51  $\pm$  0.65	&96.01  $\pm$  0.79&89.90  $\pm$  2.05&93.15  $\pm$  0.89\\
				\hline
				\multirow{3}{*}{Semeion}&4& 93.02  $\pm$  1.67		& 90.44  $\pm$  1.93		& \textbf{93.28  $\pm$  2.99}	&81.48  $\pm$  2.82&79.69  $\pm$  2.41&86.54  $\pm$  3.98\\
				&6& \textbf{86.77  $\pm$  1.35}		& 83.16  $\pm$  0.64		& 85.57  $\pm$  0.84	&75.75  $\pm$  1.98&55.98  $\pm$  8.04&79.82  $\pm$  0.99\\
				&8& 82.11  $\pm$  1.59		& \textbf{82.24  $\pm$  1.67}		& 82.11  $\pm$  0.66	&69.79  $\pm$  3.70&46.25  $\pm$  5.07&67.98  $\pm$  1.66\\
				\hline
				\multirow{3}{*}{SVHN}&4&\textbf{85.92  $\pm$  2.75}&56.58  $\pm$  16.02&71.03  $\pm$  2.70&74.11  $\pm$  4.62&22.52  $\pm$  0.40&79.77  $\pm$  0.97\\
				&6&\textbf{86.40  $\pm$  1.90}&80.98  $\pm$  14.10&50.05  $\pm$  18.30&65.52  $\pm$  5.64&13.20  $\pm$  0.03&79.90  $\pm$  0.72\\
				&8&\textbf{86.77  $\pm$  2.11}&31.43  $\pm$  17.91&33.75  $\pm$  14.40&64.39  $\pm$  4.79&8.81  $\pm$  0.01&79.34  $\pm$  0.68\\
				\hline
				\multirow{3}{*}{USPS}&4& 94.97  $\pm$  0.87		& \textbf{95.85  $\pm$  0.22}		& 95.24  $\pm$  0.22	&95.06  $\pm$  0.67&79.33  $\pm$  5.70&87.76  $\pm$  5.51\\
				&6& 92.65  $\pm$  0.67		& 93.20  $\pm$  0.41		& 92.21  $\pm$  0.92	&\textbf{93.72  $\pm$  0.92}&80.74  $\pm$  5.53&90.45  $\pm$  0.66\\
				&8& \textbf{91.33  $\pm$  0.57}		& 90.68  $\pm$  0.43		& 90.04  $\pm$  0.47	&90.59  $\pm$  1.08&75.29  $\pm$  4.18&90.50  $\pm$  1.74\\
				\hline
				{WaveForm1}&3& \textbf{83.38  $\pm$  1.32}		& 82.45  $\pm$  1.61		& 81.75  $\pm$  0.31	&71.75  $\pm$  3.02&26.91  $\pm$  4.79&64.07  $\pm$  5.60\\
				\hline
			\end{tabular}
		\end{center}
	\end{table*}
	\section{Experiments}\label{sec:5}
	In this section, we present an extensive empirical study of the proposed cost-sensitive MPU method (CSMPU). We first describe the benchmark datasets and experimental setup, then specify the model architecture and baselines for comparison, and finally report results under varying class-prior settings. Our code and experiment scripts are available online.\footnote{\url{https://github.com/EricZhM/Multi-positive-and-unlabeled-learning}}
	
	\subsection{Datasets and Experimental Setup}
	We evaluate CSMPU on \emph{eight} widely used benchmarks spanning diverse domains, including handwritten digit recognition (MNIST, KMNIST, USPS, Semeion, SVHN, PBRHD), fashion item classification (FashionMNIST), and waveform signal analysis (Waveform-1). These datasets enable a comprehensive assessment across heterogeneous data distributions, feature spaces, and class complexities. Brief descriptions follow:
	\begin{itemize}
		\item \textbf{MNIST}~\cite{lecun2002gradient}: 70{,}000 grayscale images ($28{\times}28$) of handwritten digits (0--9); a standard image-classification and PU evaluation benchmark.
		\item \textbf{FashionMNIST}~\cite{xiao2017fashion}: 70{,}000 $28{\times}28$ grayscale images from 10 fashion categories (e.g., T-shirt, dress, sneaker); a drop-in MNIST replacement with higher semantic complexity.
		\item \textbf{USPS}~\cite{hull2002database}: 9{,}298 grayscale images ($16{\times}16$) of digits scanned from U.S.\ postal envelopes; smaller than MNIST but with high per-pixel information density.
		\item \textbf{KMNIST}~\cite{clanuwat2018deep}: 70{,}000 $28{\times}28$ images of 10 Kuzushiji (cursive Japanese) characters; evaluates robustness to non-Latin glyphs.
		\item \textbf{Semeion}~\cite{semeion_handwritten_digit_178}: 1{,}593 binary $16{\times}16$ digit images from 80 writers (each digit written twice), capturing intra-writer variability.
		\item \textbf{SVHN}~\cite{netzer2011reading}: Real-world house numbers (0--9) cropped from Google Street View; 73{,}257 train and 26{,}032 test color images ($32{\times}32$) with background clutter and illumination variation.
		\item \textbf{PBRHD} (Pen-Based Recognition of Handwritten Digits)~\cite{pen-based_recognition_of_handwritten_digits_81}: 10{,}992 digit samples captured via a pressure-sensitive tablet, represented by 16-dimensional coordinate features (dynamic traces).
		\item \textbf{Waveform-1}~\cite{waveform_database_generator_(version_1)_107}: A synthetic waveform dataset with 5{,}000 instances and 21 continuous attributes; originally 3-class, adapted for the MPU setting via class grouping.
	\end{itemize}

	Unless otherwise specified, all results are averaged over five independent runs with different random seeds; class-prior settings (e.g., $\pi_k\in\{0.2,0.5,0.8\}$) and other protocol details are reported alongside each figure/table caption to ensure plug-and-play reproducibility.
	
	For the UCI-style datasets (Pen-Based and Waveform-1), we adopt a standard $80\%\text{--}20\%$ train--test split. For image datasets (MNIST, FashionMNIST, USPS, KMNIST, Semeion, SVHN), we use their predefined training/test splits when available, and otherwise apply an $80\%\text{--}20\%$ random split.
	
	To simulate the MPU setting, we treat the first $k{-}1$ categories as labeled \emph{observed classes}, while the residual \emph{unknown} bucket (the negative meta-class $k$) is included exclusively in the unlabeled pool. To form a realistic mixture, we additionally leave half of each observed class unlabeled. This protocol matches our task focus: observed-class detection without attempting to disentangle the unknown remainder.
	
	We assess robustness under varying class priors by considering three proportions for the negative meta-class within the unlabeled pool, $\pi_k\in\{0.2,0.5,0.8\}$, thereby covering both class-imbalanced and near-balanced conditions. Each experiment is repeated five times with independently randomized splits and label assignments. We report mean accuracy $\pm$ standard deviation across runs in Tables~\ref{results}--\ref{results2}, ensuring statistical reliability.
	
	\subsection{Model Architecture}
	To accommodate the varying complexity and dimensionality of the benchmarks, we adopt a dataset-adaptive architecture strategy. For \textbf{relatively simple image tasks}—\textsc{USPS}, \textsc{MNIST}, \textsc{KMNIST}, \textsc{Pen-Based}, and \textsc{FashionMNIST}—we use a \textbf{5-layer multilayer perceptron (MLP)} preceded by a flattening layer that converts $2$D inputs into feature vectors. Each hidden layer has $300$ units with \textbf{ReLU} activations followed by \textbf{batch normalization}~\cite{ioffe2015batch}. For \textbf{higher-resolution, more complex data} such as \textsc{SVHN}, we employ a \textbf{20-layer residual network (ResNet-20)}~\cite{he2016deep} with identity skip connections for effective deep representation learning.
	
	All tabular datasets (UCI-style) are preprocessed with feature-wise min–max normalization:
	\[
	\boldsymbol{x}' \;=\;
	\frac{\boldsymbol{x}-\min_{\text{train}}(\boldsymbol{x})}
	{\max_{\text{train}}(\boldsymbol{x})-\min_{\text{train}}(\boldsymbol{x})},
	\]
	where $\min_{\text{train}}(\cdot)$ and $\max_{\text{train}}(\cdot)$ are computed \emph{only} on the training split and then applied to validation/test splits to prevent leakage.
	
	Optimization uses \textbf{Adam}~\cite{kingma2014adam} with default hyperparameters and no weight decay. We sweep learning rates
	$\{1{\times}10^{-3},\,5{\times}10^{-4},\,1{\times}10^{-5},\,1{\times}10^{-6}\}$ and report the model achieving the best validation accuracy. We initialize weights with the \textbf{He (Kaiming) normal} scheme~\cite{he2015delving} to match ReLU nonlinearities. Training runs for \textbf{100 epochs} with a \textbf{batch size of 512}, without early stopping, to ensure a uniform protocol across configurations.
	
	To quantify stability, each configuration is repeated \textbf{five} times with different random initializations (fixed data splits). We report the mean $\pm$ standard deviation of test accuracy across runs.
	
	\begin{figure*}[htbp]
		\centering
		\scriptsize
		\begin{tabular}{ccc}
			\includegraphics[width=5.5cm]{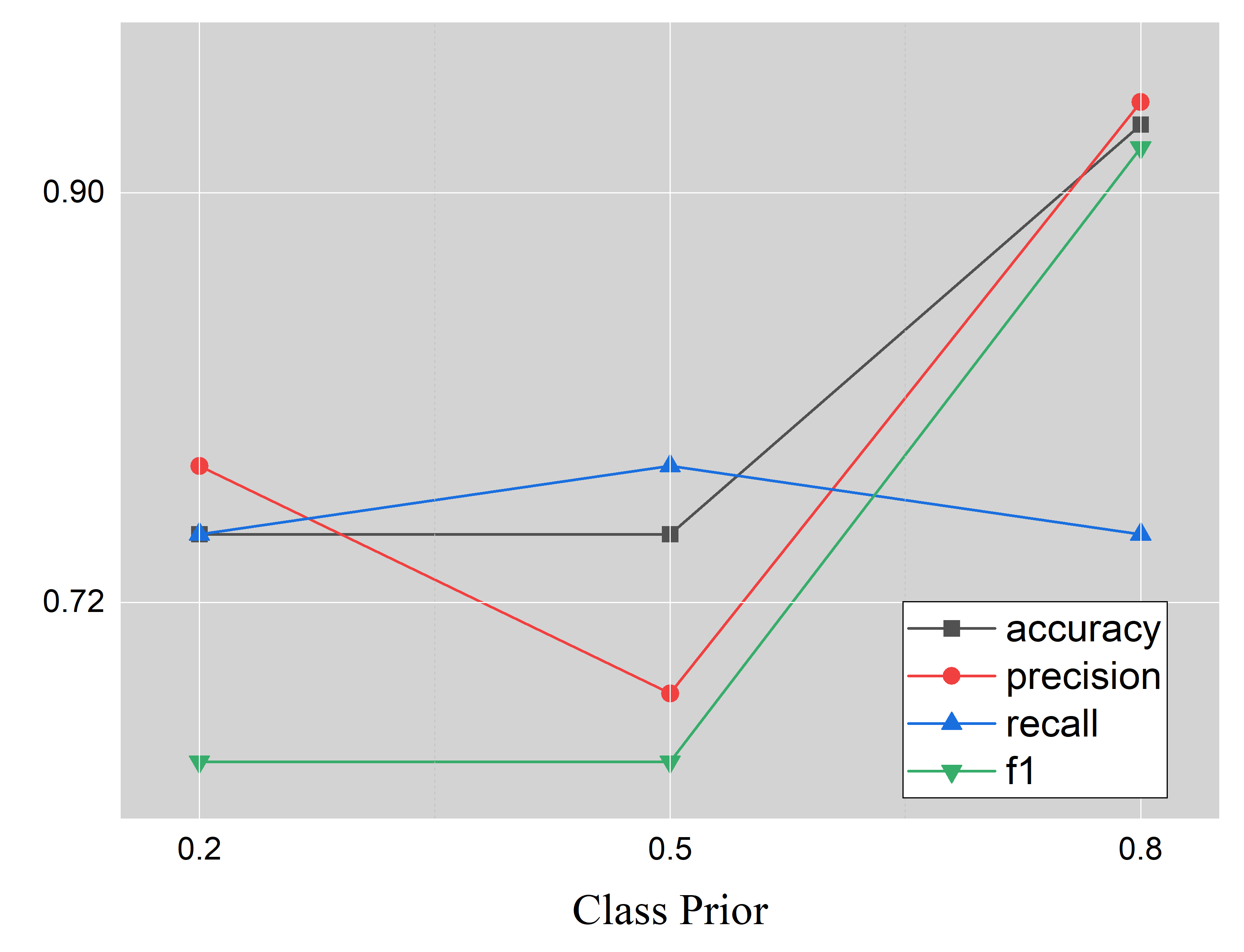} & \includegraphics[width=5.5cm]{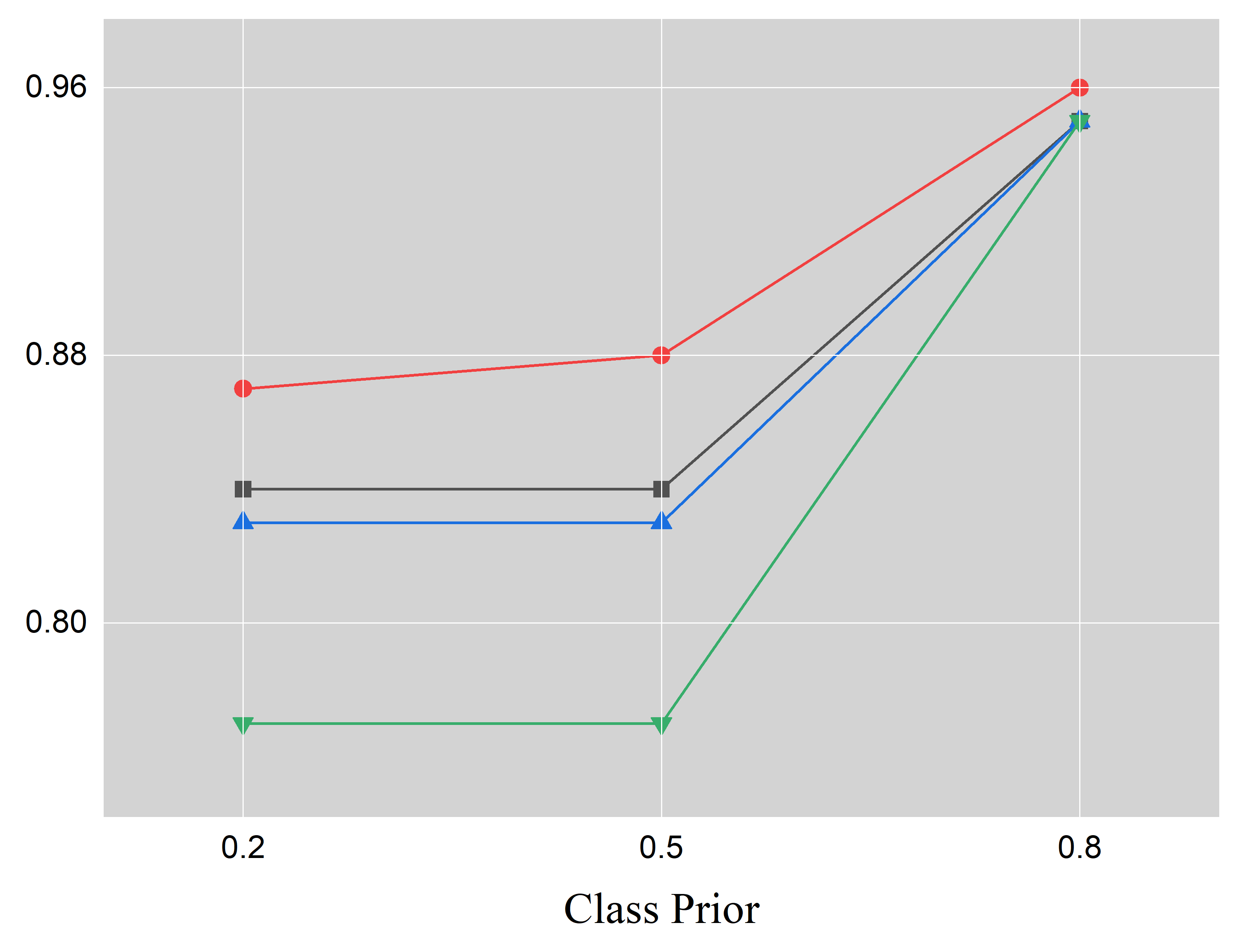} &
			\includegraphics[width=5.5cm]{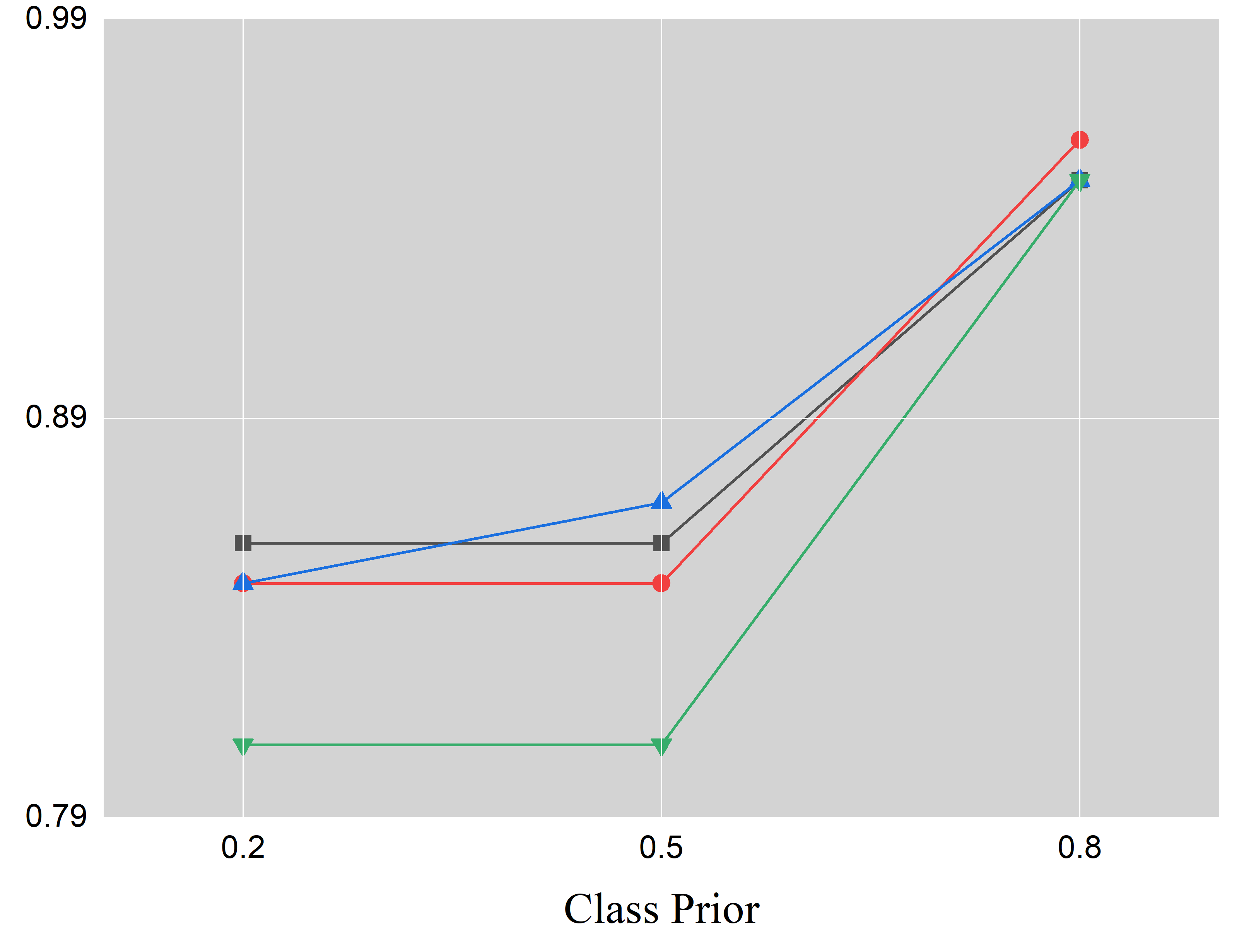} \\
			class number = 4  & class number = 6 & class number = 8 \\
		\end{tabular}
		\caption{the performance of a classification model across three different class configurations (4, 6, and 8 classes), each evaluated under varying class priors (0.2, 0.5, and 0.8). }
		\label{fig:m}
	\end{figure*}
	\begin{figure*}[!htbp]
		\centering
		\scriptsize
		\begin{tabular}{ccc}
			\includegraphics[width=5.7cm]{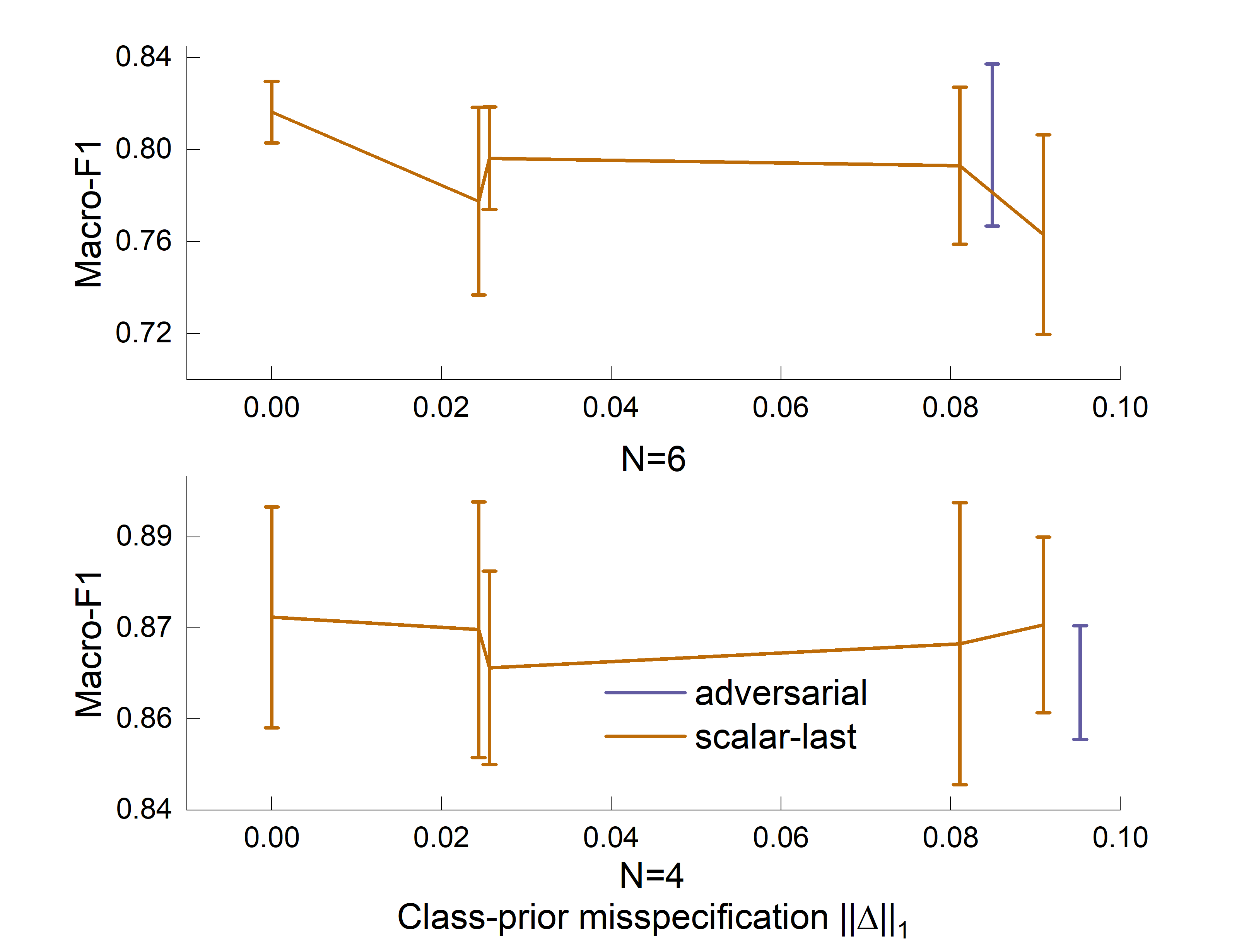} & \includegraphics[width=5.7cm]{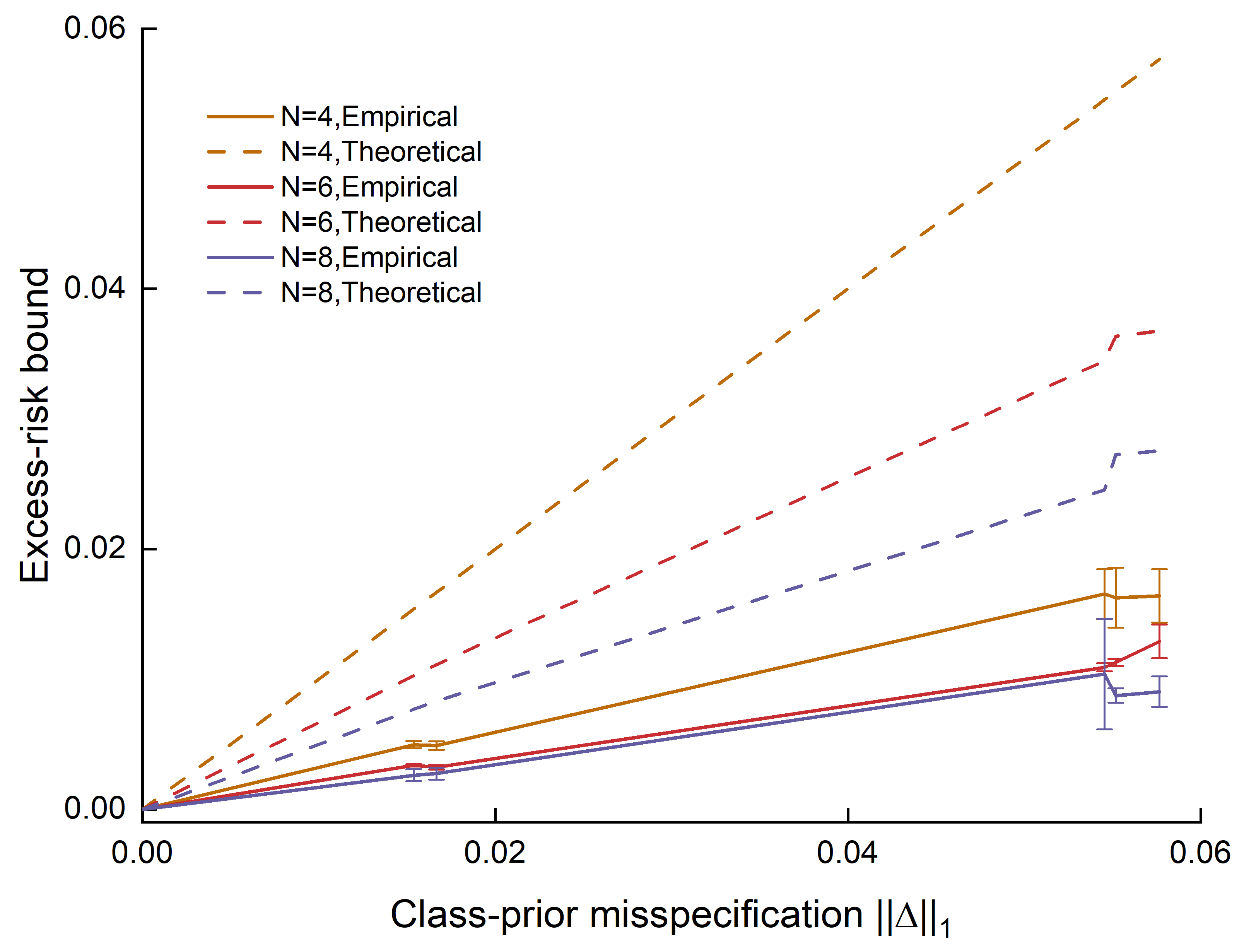} &
			\includegraphics[width=5.7cm]{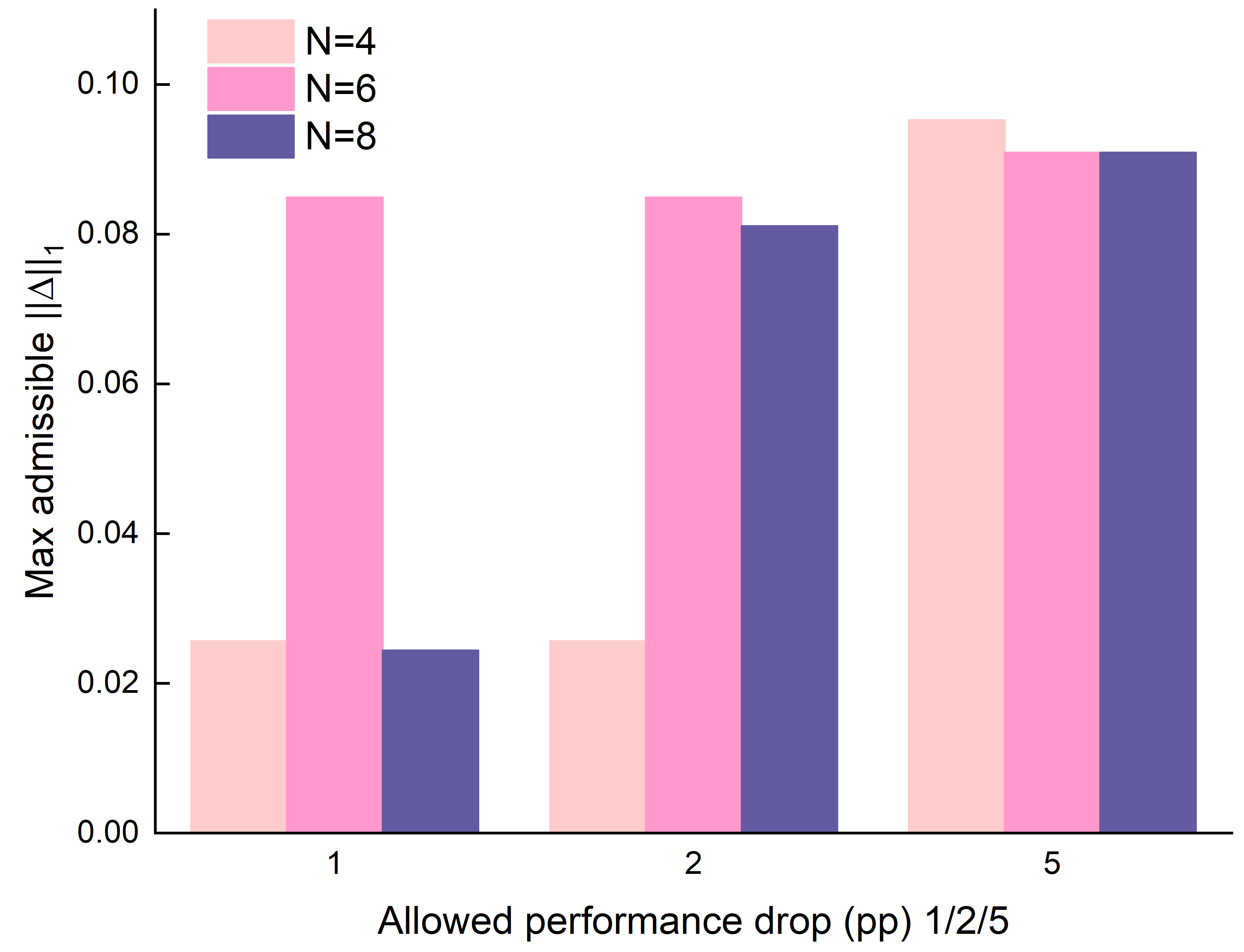} \\
			(a)  & (b) & (c) \\
		\end{tabular}
		\caption{\textbf{Effect of class-prior misspecification on performance and bounds (FashionMNIST, $\pi=0.5$).}
			(a) Macro-F1 versus misspecification magnitude $\lVert\Delta\rVert_{1}$ under two perturbation schemes (scalar-last and adversarial). We show $N{=}6$ (top) and $N{=}4$ (bottom). 
			(b) Empirical (solid) and theoretical (dashed) excess-risk bounds as functions of $\lVert\Delta\rVert_{1}$ for $N\in\{4,6,8\}$. 
			(c) Robust bandwidth: maximum admissible $\lVert\Delta\rVert_{1}$ under allowed performance drops of \{1,2,5\} percentage points.
			Here $\Delta=\hat{\pi}-\pi$ with $\sum_i \Delta_i=0$, and the empirical bound is $\sum_i \lambda_i |b_i|\,|\Delta_i|$ with $b_i=\mathbb{E}_{P_i}[\ell^{+}(g_i(X))-\ell^{-}(g_i(X))]$ and $\lambda_i=1/N$; the theory bound is $2C_{\Delta}\sum_i \lambda_i |\Delta_i|$.
			All points show mean $\pm$ SD over three independent runs.}
		\label{fig:classpriorerror}
	\end{figure*}	
	\subsection{Baseline Methods}
	\textbf{Biased-supervised (biased-super).}
	This standard supervised baseline ignores that the unlabeled pool is a mixture over all classes and naively treats \emph{all} unlabeled examples as belonging to the negative meta-class $k$. It then trains with a conventional multiclass OVR surrogate. The per-sample loss is
	\[
	\mathcal{L}\big(f(\boldsymbol{x}),y\big)
	\;=\;
	\ell\!\big(f_y(\boldsymbol{x}),+1\big)
	\;+\;
	\frac{1}{k-1}\sum_{i\neq y}\ell\!\big(f_i(\boldsymbol{x}),-1\big),
	\]
	and the empirical objective is
	\[
	\hat{\mathcal{R}}_{\mathrm{BS}}(f)
	\;=\;
	\sum_{i=1}^{k-1}\frac{1}{n_i}\sum_{j=1}^{n_i}\mathcal{L}\!\big(f(\boldsymbol{x}^{(i)}_{j}),i\big)
	\;+\;
	\frac{1}{n_k}\sum_{j=1}^{n_k}\mathcal{L}\!\big(f(\boldsymbol{x}^{(k)}_{j}),k\big),
	\]
	where $\{\boldsymbol{x}^{(i)}_{j}\}_{j=1}^{n_i}$ are labeled samples of observed class $i$ and
	$\{\boldsymbol{x}^{(k)}_{j}\}_{j=1}^{n_k}$ are \emph{unlabeled} samples that this baseline incorrectly assigns to class $k$.
	
	This approach is simple but fundamentally flawed in MPU: by mislabeling the unlabeled mixture as pure negatives, it shifts decision boundaries toward observed classes and induces systematic bias, leading to degraded accuracy and poor calibration.
	
	\textbf{\URE} (Unbiased Risk Estimator).
	This baseline constructs an estimator that is unbiased in expectation under MPU. Using the composite loss
	\[
	\tilde{\mathcal{L}}\big(f(\boldsymbol{x}),y\big)
	:= \mathcal{L}\big(f(\boldsymbol{x}),y\big)
	- \mathcal{L}\big(f(\boldsymbol{x}),k\big),
	\]
	with the OVR per-sample surrogate
	\[
	\mathcal{L}\big(f(\boldsymbol{x}),y\big)
	= \ell\!\big(f_y(\boldsymbol{x}),+1\big)
	+ \frac{1}{k-1}\sum_{i\neq y}\ell\!\big(f_i(\boldsymbol{x}),-1\big),
	\]
	the empirical objective is
	\[
	\hat{\mathcal{R}}_{\mathrm{U}}(f)
	=
	\sum_{i=1}^{k-1}\pi_i\,
	\frac{1}{n_i}\sum_{j=1}^{n_i}
	\tilde{\mathcal{L}}\!\big(f(\boldsymbol{x}^{(i)}_{j}),\,i\big)
	\;+\;
	\frac{1}{n_k}\sum_{j=1}^{n_k}
	\mathcal{L}\!\big(f(\boldsymbol{x}^{(k)}_{j}),\,k\big),
	\]
	where $\{\boldsymbol{x}^{(i)}_{j}\}_{j=1}^{n_i}$ are labeled examples of observed class $i$ and
	$\{\boldsymbol{x}^{(k)}_{j}\}_{j=1}^{n_k}$ are \emph{unlabeled} samples.
	Under mild assumptions,
	$\mathbb{E}\big[\hat{\mathcal{R}}_{\mathrm{U}}(f)\big]=\mathcal{R}(f)$,
	so the estimator is theoretically sound. In practice, however, the subtraction in
	$\tilde{\mathcal{L}}$ introduces negative sample-level components, which can yield numerically
	negative empirical risks, causing optimization instability and overfitting.
	
	\textbf{AREA}~\cite{shu2020learning}.
	AREA redesigns the risk to avoid explicit subtraction, producing an always nonnegative objective:
	\[
	\begin{split}
		\hat{\mathcal{R}}_{\mathrm{A}}(f)
		&=
		\frac{1}{n_k}\sum_{j=1}^{n_k}
		\left(
		\ell\!\big(f_k(\boldsymbol{x}^{(k)}_{j})\big)
		+ \frac{1}{k-1}\sum_{i=1}^{k-1}\ell\!\big(-f_i(\boldsymbol{x}^{(k)}_{j})\big)
		\right)
		\\
		&\quad+\;
		\frac{k}{k-1}\sum_{i=1}^{k-1}
		\frac{\pi_i}{n_i}\sum_{j=1}^{n_i}
		\left(
		\ell\!\big(f_i(\boldsymbol{x}^{(i)}_{j})\big)
		+ \ell\!\big(-f_k(\boldsymbol{x}^{(i)}_{j})\big)
		\right),
	\end{split}
	\]
	where we use the shorthand $\ell(\pm z):=\ell(z,\pm1)$.
	AREA enforces an adversarial push--pull structure (raising positive scores, suppressing negatives)
	without risky differences, which stabilizes training empirically.
	However, this structural change introduces a nonvanishing constant in the generalization analysis,
	so the bound contains a term that does not disappear as data grow, limiting the asymptotic rate.
	\begin{figure*}[htbp]
		\centering
		\scriptsize
		\begin{tabular}{cc}
			\includegraphics[width=8cm]{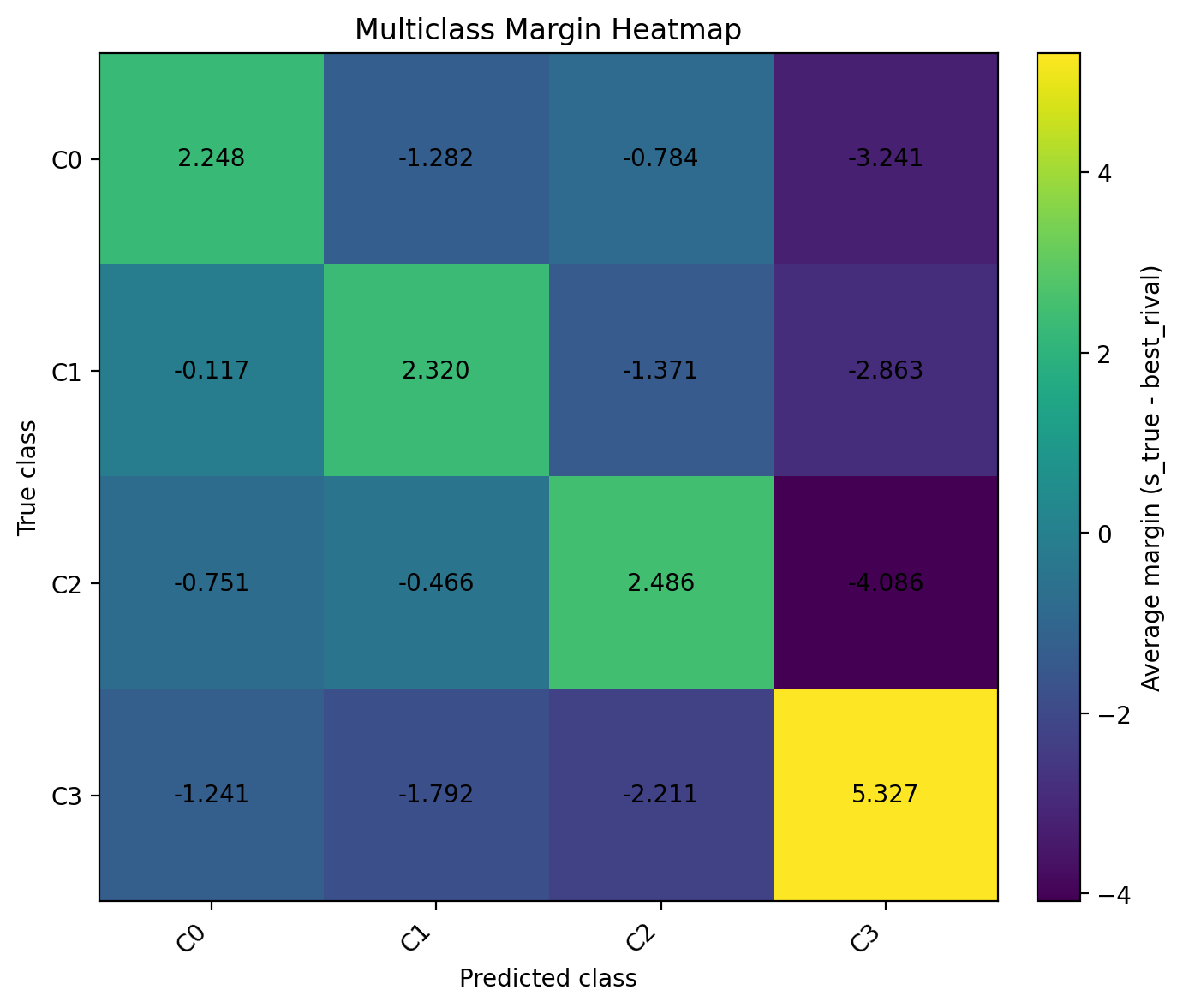} &
			\includegraphics[width=8cm]{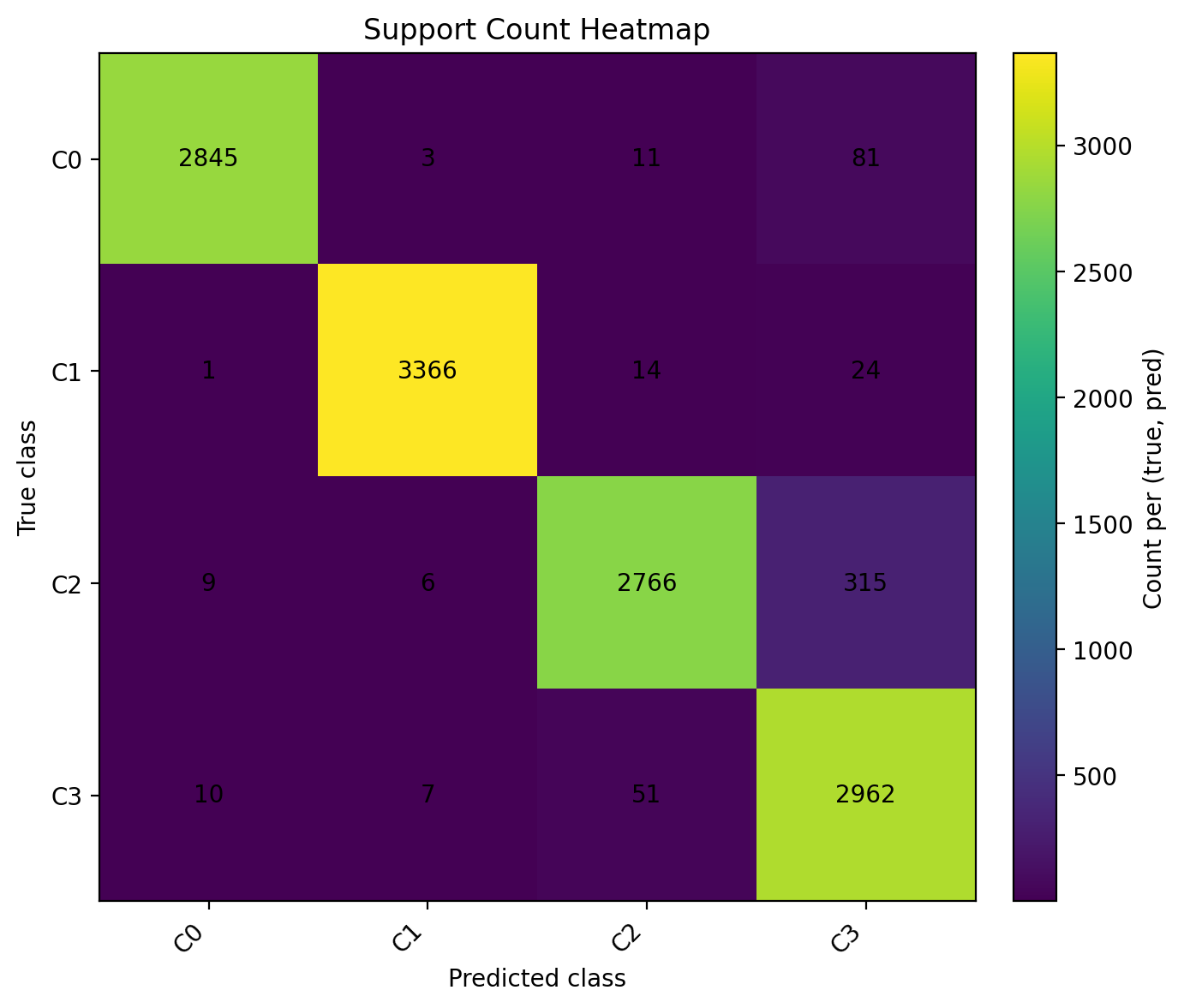} \\
			Multiclass margin heatmap & Support count heatmap \\
		\end{tabular}
		\caption{Diagnostics on \textbf{MNIST} under the MPU setting with the \(k\)-th class prior set to \textbf{0.2}. 
			Left: average margin for each (true, predicted) pair, where positive values indicate that the true class outscores its strongest rival on average; blank cells denote no support. 
			Right: support counts for the same pairs, with diagonals indicating correct predictions and off–diagonals showing confusions. 
			The two views together reveal where the model is confident versus fragile and how many samples underpin each estimate.}
		\label{fig:mnist-margin-support}
	\end{figure*}
	\begin{table*}[t]
		\centering
		\small
		\caption{Performance and constant--sum diagnostics across loss variants on \textbf{FashionMNIST} ($k{=}4$, $\pi_k{=}0.5$, mode=\texttt{CSMPU-ABS}). Reported are macro F1 and Acc ($\text{mean}\pm\text{std}$ over runs) and the numerical check of the constant--sum property via $\max$ and $p_{99}$ of $\lvert\ell(z)+\ell(-z)-C\rvert$ on a dense margin grid. Variants with near-zero values strictly satisfy the assumption; \texttt{sym} denotes the symmetrized-and-clipped version enforcing $C{=}1$, while \texttt{raw} is the original form. The probability-style \texttt{sigmoid\_prob} is distinct from the logistic loss.}
		\label{tab:loss_variants_fmnist_k4}
		\begin{tabular}{lrlllll}
			\hline
			los\_fun & gamma & sym\_clip & F1 (mean$\pm$std) & Acc (mean$\pm$std) & Const-sum max (mean) & Const-sum p99 (mean) \\
			\hline
			hinge         & 1.0 & raw & 0.86 $\pm$ 0.03 & 0.86 $\pm$ 0.03 & 10.00 &  9.90 \\
			hinge         & 1.0 & sym & 0.83 $\pm$ 0.05 & 0.83 $\pm$ 0.04 &  0.00 &  0.00 \\
			logistic      & 1.0 & raw & 0.72 $\pm$ 0.03 & 0.72 $\pm$ 0.03 &  9.00 &  8.90 \\
			ramp          & 1.0 & raw & 0.83 $\pm$ 0.06 & 0.83 $\pm$ 0.06 &  1.00 &  0.90 \\
			ramp          & 1.0 & sym & 0.83 $\pm$ 0.03 & 0.84 $\pm$ 0.02 &  0.00 &  0.00 \\
			sigmoid\_prob & 1.0 & raw & 0.85 $\pm$ 0.02 & 0.85 $\pm$ 0.02 & $1.19\times10^{-7}$ & $1.19\times10^{-7}$ \\
			sigmoid\_prob & 2.0 & raw & 0.70 $\pm$ 0.13 & 0.75 $\pm$ 0.09 & $1.19\times10^{-7}$ & $1.19\times10^{-7}$ \\
			tanh\_smooth  & 1.0 & raw & 0.83 $\pm$ 0.03 & 0.84 $\pm$ 0.02 &  0.00 &  0.00 \\
			unhinged      & 1.0 & raw & 0.65 $\pm$ 0.23 & 0.70 $\pm$ 0.19 & $2.38\times10^{-7}$ & $2.38\times10^{-7}$ \\
			\hline
		\end{tabular}
	\end{table*}
	
	\begin{table*}[t]
		\centering
		\small
		\caption{Macro-F1 on \textbf{FashionMNIST} ($ k{=}4$, $\pi_k{=}0.8$). Values are mean$\pm$std over $n$ seeds; 95\% CI in brackets. $p_{\text{Holm}}$ from paired Wilcoxon vs.~\textsc{Ours}; $\Delta$ is Cliff's delta (effect size).Paired two-sided Wilcoxon; Holm correction across rows in this table. Cliff's $\Delta{<}0$ favors the baseline; $\Delta{>}0$ favors the method. CI via $t$-interval; Time is median$\pm$MAD.}
		\label{tab:fmnist_k4_pi08}
		\begin{tabular}{lccc}
			\hline
			Method & Macro\_F1 (\%) $\uparrow$ & Time (min) $\downarrow$ & vs. Ours $[p_\text{Holm},\,\Delta]$ \\
			\hline
			AREA & 79.6$\pm$6.8 [71.1, 88.1] & 0.5$\pm$0.0 & [1,\,-1.00] \\
			URE-OVR & 27.5$\pm$12.5 [12.0, 43.0] & 0.6$\pm$0.0 & [1,\,-1.00] \\
			\textbf{Ours} & 89.9$\pm$2.1 [87.3, 92.5] & 0.6$\pm$0.0 & -- \\
			\hline
		\end{tabular}
	\end{table*}
	
	Table~\ref{tab:loss_variants_fmnist_k4} compares binary surrogate losses with respect to the constant–sum condition
	$\ell(z)+\ell(-z)=C$ (with $z=yf(\boldsymbol{x})$, $y\in\{\pm1\}$) and their empirical performance.
	Losses that \emph{exactly} satisfy the identity include
	\emph{unhinged} $\ell_{\mathrm{unh}}(z)=\tfrac{1-z}{2}$ ($C=1$),
	the \emph{probability-style sigmoid} $\ell_{\sigma}(z)=\sigma(-\gamma z)=\tfrac{1}{1+e^{\gamma z}}$ ($C=1$; note this is \emph{not} the logistic log-loss),
	and the \emph{tanh-smooth} loss $\ell_{\tanh}(z)=\tfrac{1-\tanh(\gamma z)}{2}$ ($C=1$).
	By contrast, the standard \emph{hinge} $\ell_{\mathrm{hinge}}(z)=\max\{0,1-z\}$,
	\emph{ramp} $\ell_{\mathrm{ramp}}(z)=\min\{1,\max\{0,1-z\}\}$,
	and the logistic log-loss $\ell_{\mathrm{log}}(z)=\log(1+e^{-\gamma z})$ do \emph{not} satisfy the condition.
	
	To render hinge/ramp theoretically admissible, we use the symmetrized–and–clipped variant
	$\ell_{\mathrm{sym}}(z)=\tfrac{\ell(z)-\ell(-z)}{2}+\tfrac{C}{2}$, clipped to $[0,C]$ (we set $C=1$),
	which guarantees $\ell_{\mathrm{sym}}(z)+\ell_{\mathrm{sym}}(-z)=C$.
	Near-zero values in the columns ``Const-sum max / p99'' indicate the identity holds:
	\texttt{sigmoid\_prob}, \texttt{tanh\_smooth}, \texttt{unhinged}, and the \texttt{sym} versions of \texttt{hinge}/\texttt{ramp} show near-zero errors,
	whereas the \texttt{raw} versions of \texttt{hinge}/\texttt{logistic}/\texttt{ramp} deviate substantially and are unsuitable for unbiased cancellation.
	In terms of accuracy, the constant--sum-satisfying \texttt{sigmoid\_prob} with $\gamma{=}1$ achieves the best macro-F1 under this setting;
	converting \texttt{hinge} from \texttt{raw} to \texttt{sym} trades a small amount of accuracy for theoretical validity,
	and overly large $\gamma$ for \texttt{sigmoid\_prob} degrades performance.
	\subsection{Results and Analysis}
	Across \emph{eight} benchmark datasets and three negative–meta-class priors $\pi_k\!\in\!\{0.2,0.5,0.8\}$, the proposed CSMPU framework demonstrates strong and consistent performance. As shown in Figure~\ref{fig:compared}, CSMPU exhibits smooth convergence of both training and test accuracy on MNIST, FashionMNIST, and KMNIST, without signs of severe overfitting. This stability contrasts with MPU baselines such as \URE\ and AREA, which often display fluctuating trajectories and widening train–test gaps. The outer nonnegativity correction in CSMPU prevents negative empirical risks and contributes to robust generalization, even with deep encoders.
	
	Tables~\ref{results}--\ref{results2} further show that CSMPU consistently outperforms or matches the strongest baselines across diverse domains. On high-dimensional image datasets (MNIST, FashionMNIST, KMNIST, SVHN), the advantage is most pronounced when the negative prior is small—precisely when the unlabeled pool is most ambiguous. For example, on MNIST with six classes and $\pi_k=0.2$, CSMPU attains $92.73\%\pm0.83$, substantially higher than AREA at $79.03\%$. Similar trends appear on SVHN, where CSMPU reaches $86.40\%\pm1.90$ under $\pi_k=0.8$ while competing estimators degrade. Even on lower-dimensional or synthetic data (Waveform-1), where many baselines collapse toward chance under $\pi_k=0.2$, CSMPU sustains $>80\%$ accuracy, reflecting resilience under heavy imbalance and noise.
	
	Figure~\ref{fig:m} reports \emph{CSMPU alone} across class counts
	$K\!\in\!\{4,6,8\}$: all four metrics generally increase as the class prior
	$\pi_k$ grows. The sensitivity to $\pi_k$ is stronger at smaller $K$ ($4$ or
	$6$), whereas for $K{=}8$ the curves still improve but flatten, indicating milder
	gains and slightly more conservative recall at low priors.
	
	Figure~\ref{fig:mnist-margin-support} highlights two diagnostics of the trained CSMPU model. 
	First, the \emph{support-count heatmap} shows strong diagonals with substantial sample support for every class (C0: $2845$, C1: $3366$, C2: $2766$, C3: $2962$), indicating that predictions concentrate on the correct class rather than off-diagonals. 
	Second, the \emph{margin heatmap} reports positive class-wise mean margins (C0: $2.248$, C1: $2.320$, C2: $2.486$, C3: $5.327$), meaning that, on average, the true-class score exceeds its strongest rival by a clear gap. 
	Together, high on-diagonal counts and positive mean margins point to confident, well-separated decisions across classes in this experiment; notably, class~C3 attains the largest mean margin ($5.327$), while the others remain in a similar positive range ($2.248$–$2.486$).

	We quantify sensitivity to prior misspecification by perturbing the assumed prior $\hat{\pi}$ as $\Delta=\hat{\pi}-\pi$ with $\sum_i \Delta_i=0$, under two schemes (\emph{scalar-last} and \emph{adversarial}) and for $N\!\in\!\{4,6,8\}$.
	As shown in Fig.~\ref{fig:classpriorerror}(a), Macro-F1 is essentially flat for small $\|\Delta\|_{1}$ and degrades smoothly as $\|\Delta\|_{1}$ grows; the adversarial scheme is most harmful yet yields only a modest drop over the tested range.
	Fig.~\ref{fig:classpriorerror}(b) compares the empirical excess-risk proxy $\sum_i \lambda_i |b_i|\,|\Delta_i|$ (solid) with the theoretical bound $2C_{\Delta}\sum_i \lambda_i |\Delta_i|$ (dashed), where $b_i=\mathbb{E}_{\boldsymbol{X}\sim P_i}\!\big[\ell^{+}(f_i(\boldsymbol{X}))-\ell^{-}(f_i(\boldsymbol{X}))\big]$ and $\lambda_i=1/N$.
	The empirical curve grows nearly linearly with $\|\Delta\|_{1}$ and consistently lies below the theoretical envelope, matching our analysis up to constants.
	Finally, Fig.~\ref{fig:classpriorerror}(c) reports the \emph{robust bandwidth}—the maximum admissible $\|\Delta\|_{1}$ under allowed performance drops of $\{1,2,5\}$ percentage points—indicating practically useful tolerance to prior misspecification across $N\in\{4,6,8\}$.
	All points denote mean $\pm$ standard deviation over three independent runs.

	Table~\ref{tab:fmnist_k4_pi08} reports Macro-F1 on \textsc{FashionMNIST} with $k{=}4$ and $\pi_k{=}0.8$, listing mean$\pm$std over $n$ seeds with 95\% CIs in brackets and, in the last column, paired Wilcoxon tests (Holm-adjusted) against \textsc{Ours} plus Cliff’s~$\Delta$ (effect size). \textsc{Ours} achieves the best Macro-F1 (89.9$\pm$2.1; CI [87.3, 92.5]), compared with \textsc{AREA} (79.6$\pm$6.8; [71.1, 88.1]) and \textsc{URE-OVR} (27.5$\pm$12.5; [12.0, 43.0]); the effect sizes indicate complete separation ($\Delta\!\approx\!-1.00$), the narrower CI suggests greater stability, and runtime is comparable across methods ($\approx$0.5--0.6 min), showing that the gains are not achieved at additional computational cost.
	\section{Conclusion}\label{sec:6}
	We presented \textbf{CSMPU}, a cost-sensitive OVR framework for MPU learning targeted at \emph{observed-class detection}. The method couples a class-dependent OVR objective with an unbiased MPU risk and applies a hard nonnegativity correction at the outer (aggregated) level, yielding a simple, modular training criterion that requires no supervision beyond labeled positives and integrates cleanly with modern neural encoders.
	
	On the theory side, we established a Rademacher-based generalization bound for the corrected objective and quantified the effect of class-prior misspecification via explicit bias bounds. Empirically, across eight benchmarks and a range of negative–meta-class priors, CSMPU delivers consistently higher accuracy and stability than representative MPU baselines, with smooth optimization and strong robustness when the unlabeled pool is most ambiguous. Taken together, the results show that enforcing cost-sensitive per-class targets within an unbiased, nonnegative MPU risk leads to a practical and reliable solution for multi-class PU learning.
	\bibliographystyle{ieeetr}
	\bibliography{ref}

\begin{thebibliography}{10}

\bibitem{zhou2018brief}
Z.-H. Zhou, ``A brief introduction to weakly supervised learning,'' {\em
  National Science Review}, vol.~5, no.~1, pp.~44--53, 2018.

\bibitem{10847040}
A.~D~A, M.~S. Shekar, A.~Bharadwaj, N.~Vineeth, and M.~L. Neelima, ``Deep
  learning in medical image analysis: A survey,'' in {\em 2024 International
  Conference on Innovation and Novelty in Engineering and Technology (INNOVA)},
  vol.~I, pp.~1--5, 2024.

\bibitem{10039501}
L.~Sui, C.-L. Zhang, and J.~Wu, ``Salvage of supervision in weakly supervised
  object detection and segmentation,'' {\em IEEE Transactions on Pattern
  Analysis and Machine Intelligence}, vol.~45, no.~8, pp.~10394--10408, 2023.

\bibitem{9444588}
T.~Jiang, W.~Xie, Y.~Li, J.~Lei, and Q.~Du, ``Weakly supervised discriminative
  learning with spectral constrained generative adversarial network for
  hyperspectral anomaly detection,'' {\em IEEE Transactions on Neural Networks
  and Learning Systems}, vol.~33, no.~11, pp.~6504--6517, 2022.

\bibitem{10172241}
J.~Yu, H.~Oh, M.~Kim, and J.~Kim, ``Weakly supervised contrastive learning for
  unsupervised vehicle reidentification,'' {\em IEEE Transactions on Neural
  Networks and Learning Systems}, vol.~35, no.~11, pp.~15543--15553, 2024.

\bibitem{10472151}
L.~Xu, M.~Bennamoun, F.~Boussaid, W.~Ouyang, F.~Sohel, and D.~Xu, ``Auxiliary
  tasks enhanced dual-affinity learning for weakly supervised semantic
  segmentation,'' {\em IEEE Transactions on Neural Networks and Learning
  Systems}, vol.~36, no.~3, pp.~5082--5096, 2025.

\bibitem{bekker2020learning}
J.~Bekker and J.~Davis, ``Learning from positive and unlabeled data: A
  survey,'' {\em Machine Learning}, vol.~109, no.~4, pp.~719--760, 2020.

\bibitem{jiang2023positive}
Y.~Jiang, Q.~Xu, Y.~Zhao, Z.~Yang, P.~Wen, X.~Cao, and Q.~Huang,
  ``Positive-unlabeled learning with label distribution alignment,'' {\em IEEE
  Transactions on Pattern Analysis and Machine Intelligence}, vol.~45, no.~12,
  pp.~15345--15363, 2023.

\bibitem{10870373}
B.~Yuan, C.~Gong, D.~Tao, and J.~Yang, ``Weighted contrastive learning with
  hard negative mining for positive and unlabeled learning,'' {\em IEEE
  Transactions on Neural Networks and Learning Systems}, vol.~36, no.~6,
  pp.~10515--10529, 2025.

\bibitem{chang2021positive}
S.~Chang, B.~Du, and L.~Zhang, ``Positive unlabeled learning with class-prior
  approximation,'' in {\em Proceedings of the Twenty-Ninth International
  Conference on International Joint Conferences on Artificial Intelligence},
  pp.~2014--2021, 2021.

\bibitem{9583860}
M.~Qian, Y.-F. Li, and T.~Han, ``Positive-unlabeled learning-based hybrid deep
  network for intelligent fault detection,'' {\em IEEE Transactions on
  Industrial Informatics}, vol.~18, no.~7, pp.~4510--4519, 2022.

\bibitem{10792933}
P.~Kumar, F.~Moomtaheen, S.~A. Malec, J.~J. Yang, C.~G. Bologa, K.~A.
  Schneider, and Y.~e.~a. Zhu, ``Detecting opioid use disorder in health claims
  data with positive unlabeled learning,'' {\em IEEE Journal of Biomedical and
  Health Informatics}, vol.~29, no.~2, pp.~750--757, 2025.

\bibitem{ijcai2017p444}
Y.~Xu, C.~Xu, C.~Xu, and D.~Tao, ``Multi-positive and unlabeled learning,'' in
  {\em Proceedings of the Twenty-Sixth International Joint Conference on
  Artificial Intelligence, {IJCAI-17}}, pp.~3182--3188, 2017.

\bibitem{shu2020learning}
S.~Shu, Z.~Lin, Y.~Yan, and L.~Li, ``Learning from multi-class positive and
  unlabeled data,'' in {\em Proceedings of the 2020 IEEE International
  Conference on Data Mining (ICDM)}, pp.~1256--1261, IEEE, 2020.

\bibitem{fotopoulou2024review}
S.~Fotopoulou, ``A review of unsupervised learning in astronomy,'' {\em
  Astronomy and Computing}, vol.~48, p.~100851, 2024.

\bibitem{bhat2017identifying}
S.~Bhat and A.~Culotta, ``Identifying leading indicators of product recalls
  from online reviews using positive unlabeled learning and domain
  adaptation,'' in {\em Proceedings of the International AAAI Conference on Web
  and Social Media}, pp.~480--483, 2017.

\bibitem{kiryo2017positive}
R.~Kiryo, G.~Niu, M.~C. du~Plessis, and M.~Sugiyama, ``Positive-unlabeled
  learning with non-negative risk estimator,'' in {\em Proceedings of the 31st
  International Conference on Neural Information Processing Systems}, NIPS'17,
  (Red Hook, NY, USA), p.~1674–1684, Curran Associates Inc., 2017.

\bibitem{elkan2008learning}
C.~Elkan and K.~Noto, ``Learning classifiers from only positive and unlabeled
  data,'' in {\em Proceedings of the 14th ACM SIGKDD international conference
  on Knowledge discovery and data mining}, pp.~213--220, 2008.

\bibitem{du2014analysis}
M.~C. du~Plessis, G.~Niu, and M.~Sugiyama, ``Analysis of learning from positive
  and unlabeled data,'' in {\em Advances in Neural Information Processing
  Systems} (Z.~Ghahramani, M.~Welling, C.~Cortes, N.~Lawrence, and
  K.~Weinberger, eds.), vol.~27, Curran Associates, Inc., 2014.

\bibitem{elkan2001foundations}
C.~Elkan, ``The foundations of cost-sensitive learning,'' in {\em Proceedings
  of the International Joint Conference on Artificial Intelligence (IJCAI)},
  pp.~973--978, 2001.

\bibitem{sun2007cost}
Z.-H. Sun, A.~K. Wong, and M.~S. Kamel, ``Cost-sensitive boosting for
  classification of imbalanced data,'' {\em Pattern Recognition}, vol.~40,
  no.~12, pp.~3358--3378, 2007.

\bibitem{zhou2010multi}
Z.-H. Zhou and X.-Y. Liu, ``Multi-class cost-sensitive learning with
  applications to credit scoring and intrusion detection,'' in {\em Proceedings
  of the 2010 IEEE International Conference on Data Mining}, pp.~619--628,
  IEEE, 2010.

\bibitem{khan2017cost}
S.~H. Khan, M.~Hayat, M.~Bennamoun, F.~A. Sohel, and R.~Togneri,
  ``Cost-sensitive learning of deep feature representations from imbalanced
  data,'' {\em IEEE Transactions on Neural Networks and Learning Systems},
  vol.~29, no.~8, pp.~3573--3587, 2017.

\bibitem{o2008cost}
D.~B. O'Brien, M.~R. Gupta, and R.~M. Gray, ``Cost-sensitive multi-class
  classification from probability estimates,'' in {\em Proceedings of the 25th
  International Conference on Machine Learning}, pp.~712--719, 2008.

\bibitem{JMLR:v23:21-0946}
S.~Wu, T.~Liu, B.~Han, J.~Yu, G.~Niu, and M.~Sugiyama, ``Learning from noisy
  pairwise similarity and unlabeled data,'' {\em Journal of Machine Learning
  Research}, vol.~23, no.~307, pp.~1--34, 2022.

\bibitem{lu2018minimal}
N.~Lu, G.~Niu, A.~K. Menon, and M.~Sugiyama, ``On the minimal supervision for
  training any binary classifier from only unlabeled data,'' in {\em 7th
  International Conference on Learning Representations, {ICLR} 2019, New
  Orleans, LA, USA, May 6-9, 2019}, OpenReview.net, 2019.

\bibitem{9878485}
Y.~Zhao, Q.~Xu, Y.~Jiang, P.~Wen, and Q.~Huang, ``Dist-pu: Positive-unlabeled
  learning from a label distribution perspective,'' in {\em 2022 IEEE/CVF
  Conference on Computer Vision and Pattern Recognition (CVPR)},
  pp.~14441--14450, 2022.

\bibitem{Blanchard2010SSND}
G.~Blanchard, G.~Lee, and C.~Scott, ``Semi-supervised novelty detection,'' {\em
  Journal of Machine Learning Research}, vol.~11, no.~99, pp.~2973--3009, 2010.

\bibitem{duPlessis2017ClassPriorPU}
M.~C. du~Plessis, G.~Niu, and M.~Sugiyama, ``Class-prior estimation for
  learning from positive and unlabeled data,'' {\em Machine Learning},
  vol.~106, no.~4, pp.~463--492, 2017.
\newblock Issue date: April 2017; Published online: 2016-11-14.

\bibitem{maximov2018rademacher}
Y.~Maximov, M.-R. Amini, and Z.~Harchaoui, ``Rademacher complexity bounds for a
  penalized multi-class semi-supervised algorithm,'' {\em Journal of Artificial
  Intelligence Research}, vol.~61, pp.~761--786, 2018.

\bibitem{musayeva2019rademacher}
K.~Musayeva, F.~Lauer, and Y.~Guermeur, ``Rademacher complexity and
  generalization performance of multi-category margin classifiers,'' {\em
  Neurocomputing}, vol.~342, pp.~6--15, 2019.

\bibitem{lecun2002gradient}
Y.~LeCun, L.~Bottou, Y.~Bengio, and P.~Haffner, ``Gradient-based learning
  applied to document recognition,'' {\em Proceedings of the IEEE}, vol.~86,
  no.~11, pp.~2278--2324, 2002.

\bibitem{xiao2017fashion}
H.~Xiao, K.~Rasul, and R.~Vollgraf, ``Fashion-mnist: a novel image dataset for
  benchmarking machine learning algorithms,'' {\em arXiv:1708.07747}, 2017.

\bibitem{hull2002database}
J.~J. Hull, ``A database for handwritten text recognition research,'' {\em IEEE
  Transactions on Pattern Analysis and Machine Intelligence}, vol.~16, no.~5,
  pp.~550--554, 2002.

\bibitem{clanuwat2018deep}
T.~Clanuwat, M.~Bober-Irizar, A.~Kitamoto, A.~Lamb, K.~Yamamoto, and D.~Ha,
  ``Deep learning for classical japanese literature,'' {\em arXiv preprint
  arXiv:1812.01718}, 2018.

\bibitem{semeion_handwritten_digit_178}
``{Semeion Handwritten Digit}.'' UCI Machine Learning Repository, 1998.
\newblock {DOI}: 10.24432/C5SC8V.

\bibitem{netzer2011reading}
Y.~Netzer, T.~Wang, A.~Coates, A.~Bissacco, B.~Wu, A.~Y. Ng, {\em et~al.},
  ``Reading digits in natural images with unsupervised feature learning,'' in
  {\em NIPS Workshop on Deep Learning and Unsupervised Feature Learning}, p.~7,
  Granada, 2011.

\bibitem{pen-based_recognition_of_handwritten_digits_81}
E.~Alpaydin and F.~Alimoglu, ``{Pen-Based Recognition of Handwritten Digits}.''
  UCI Machine Learning Repository, 1996.
\newblock {DOI}: 10.24432/C5MG6K.

\bibitem{waveform_database_generator_(version_1)_107}
L.~Breiman and C.~Stone, ``{Waveform Database Generator (Version 1)}.'' UCI
  Machine Learning Repository, 1984.
\newblock {DOI}: https://doi.org/10.24432/C5CS3C.

\bibitem{ioffe2015batch}
S.~Ioffe and C.~Szegedy, ``Batch normalization: Accelerating deep network
  training by reducing internal covariate shift,'' in {\em International
  Conference on Machine Learning}, pp.~448--456, pmlr, 2015.

\bibitem{he2016deep}
K.~He, X.~Zhang, S.~Ren, and J.~Sun, ``Deep residual learning for image
  recognition,'' in {\em Proceedings of the IEEE Conference on Computer Vision
  and Pattern Recognition}, pp.~770--778, 2016.

\bibitem{kingma2014adam}
D.~P. Kingma and J.~Ba, ``Adam: A method for stochastic optimization,'' {\em
  arXiv preprint arXiv:1412.6980}, 2014.

\bibitem{he2015delving}
K.~He, X.~Zhang, S.~Ren, and J.~Sun, ``Delving deep into rectifiers: Surpassing
  human-level performance on imagenet classification,'' in {\em Proceedings of
  the IEEE International Conference on Computer Vision}, pp.~1026--1034, 2015.

\end{thebibliography}
\end{document}